\documentclass[10pt,twocolumn,letterpaper]{article}

\usepackage[pagenumbers]{wacv} % To force page numbers, e.g. for an arXiv version

\usepackage{multirow}
\usepackage{mathtools}
\usepackage{minted}
\usepackage{algorithm}
\usepackage{algpseudocode}
\usepackage{listings}
\usepackage[accsupp]{axessibility} 

\setminted{
    bgcolor=gray!10, % Light gray background (10% gray)
    fontsize=\small, % Use a smaller font size
    linenos,         % Add line numbers
    frame=lines,     % Add a frame around the code
    framesep=2mm,
    breaklines       % Automatically break long lines
}
\definecolor{wacvblue}{rgb}{0.21,0.49,0.74}
\usepackage[pagebackref,breaklinks,colorlinks,allcolors=wacvblue]{hyperref}

\def\wacvPaperID{2192} % *** Enter the WACV Paper ID here
\def\confName{WACV}
\def\confYear{2026}

\title{START: \underline{S}patial and \underline{T}extual Learning for Ch\underline{art} Understanding}

\author{Zhuoming Liu\textsuperscript{1\thanks{Work done during internship at Amazon AGI.}}, Xiaofeng Gao\textsuperscript{2}, Feiyang Niu\textsuperscript{2}, Qiaozi Gao\textsuperscript{2}, Liu Liu\textsuperscript{3}, Robinson Piramuthu\textsuperscript{2}\\ 
{\textsuperscript{1}University of Wisconsin-Madison \ \ 
\textsuperscript{2}Amazon AGI \ \
\textsuperscript{3}MIT \ \ 
}}

\begin{document}
\maketitle
\begin{abstract}
Chart understanding is crucial for deploying multimodal large language models (MLLMs) in real-world scenarios such as analyzing scientific papers and technical reports. Unlike natural images, charts pair a structured visual layout (spatial property) with an underlying data representation (textual property) — grasping both is essential for precise, fine-grained chart reasoning.
Motivated by this observation, we propose START, the Spatial and Textual learning for chART understanding. Specifically, we introduce (i) chart-element grounding and (ii) chart-to-code generation to strengthen an MLLM’s understanding of both chart visual layout and data details. 
To facilitate spatial and textual learning, we propose the START-Dataset generated with a novel data-generation pipeline that first leverages an MLLM to translate real chart images into executable chart code, recovering the underlying data representation while preserving the visual distribution of real-world charts. We then evolve the code with a Large Language Model (LLM) to ascertain the positions of chart elements that capture the chart's visual structure, addressing challenges that existing methods cannot handle. To evaluate a model’s ability to understand chart spatial structures, we propose the Chart Spatial understanding Benchmark (CS-Bench), filling a critical gap in comprehensive chart understanding evaluation.
Leveraging spatial and textual learning, START delivers consistent gains across model sizes and benchmarks over the base models and surpasses prior state-of-the-art by a clear margin. Code, data and models will be publicly available.

\end{abstract}    
\section{Introduction}
\label{sec:intro}
% the overview of the method 

\begin{figure}[t!]
    \centering
    \includegraphics[width=1.0\linewidth]{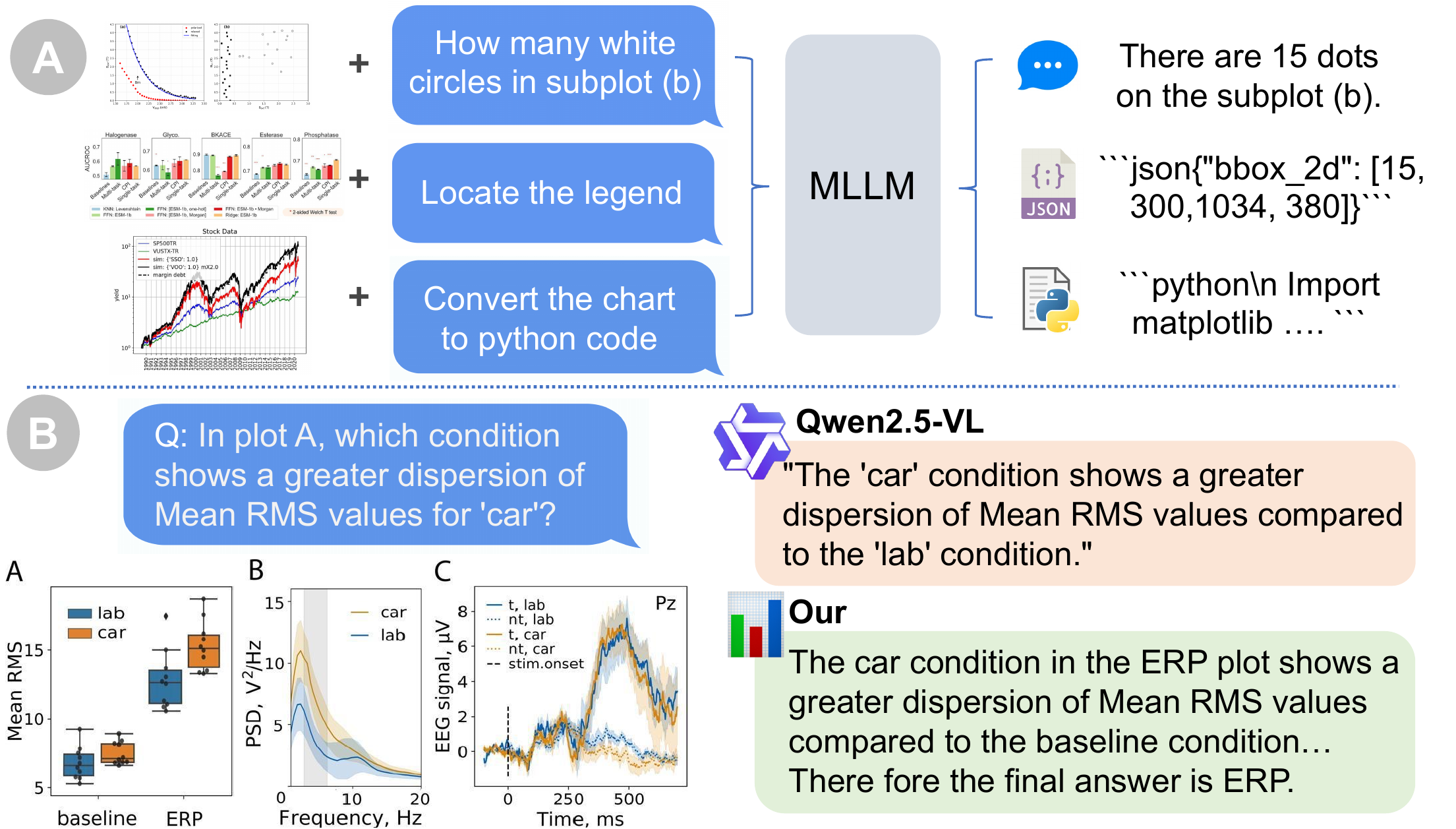}
    \vspace{-1em}
    \caption{A: the overview of the START, which leverages spatial and textual learning for chart understanding. B: challenging question sample from the CharXiv~\cite{wang2024charxiv}. Answering the question requires chart element grounding and step-by-step reasoning.}
    \label{fig:thumbnail}
    \vspace{-1.5em}
\end{figure}

% 1. MLLM and why chart understanding is important?
The rapid advancement of multimodal large language models (MLLMs) has opened new frontiers in artificial intelligence, enabling processing and reasoning across text, images, and other modalities simultaneously. As their capabilities continue to grow, successful deployment in real-world applications increasingly hinges on their ability to accurately understand and interpret complex visual information. Among various types of visual content, charts represent a particularly challenging yet essential domain for MLLM, especially in real-world scenarios such as analyzing scientific papers, technical reports, and financial documents.

% 3. Chat understanding is challenging, for example of the chart question (require grounding and details), the mistake made by the existing model, the leaderboard?
However, despite significant progress in general multimodal understanding, current MLLMs often struggle with understanding complicated visual structure and details in the charts. 
Even the best vision reasoning model OpenAI o3~\cite{o3openai} still lags behind the human level understanding to the charts~\cite{wang2024charxiv}.
Figure~\ref{fig:thumbnail}-B shows a sample question from CharXiv~\cite{wang2024charxiv}, which requires a step-by-step reasoning and chart element grounding based on the instruction in the question.
The Qwen2.5-VL~\cite{bai2025qwen2}, one of the best open-source MLLMs, makes mistakes given that it does not ground the "condition" to the x-axis correctly, justifying the difficulties of the chart's understanding.

% 2. What is a chart? What is the property of the chart
Unlike natural images that primarily convey semantic content through objects and scenes, charts are artificial visual input that pair a structured spatial layout with an underlying textual data representation.
They typically include subplots, titles, legends, and axes, and are instantiated from data sources—such as tables—or rendered by code (e.g., Python~\cite{yang2025effective, he2024distill}). 
% As a result, effective chart understanding demands integrated reasoning over both the visual structure and the textual content to achieve precise and fine-grained interpretation.
% 4. our research question?
% (should change to spatial understanding and text understanding, you also need to stress the existing data does not help complex reasoning due to the over-simplied visual appearances. Chart element location strategies could in ChartQA or PlotQA could not be applied to chart image with multiple subplot due to the structural complexity, or due to the svg files missing)
Motivated by the properties of the chart, we raise two research questions: \textit{1. Can explicitly learning the spatial structure of the chart and recovering the textual details in the chart help the chart understanding?} 
and 
% \textit{2. Can we construct a dataset to facilitate the comprehensive chart understanding, which has the chart image that reflects the data distribution in real life, also comes with the visual layout data and the underlying data representation?}
\textit{2. To facilitate spatial and textual learning, how should we construct the dataset?}

% 5. what we proposed (the method)
To address these questions, we propose \textbf{START}, \underline{S}patial and \underline{T}extual learning ch\underline{ART} understanding, as shown in Figure~\ref{fig:thumbnail}-A.
Specifically, we formalize the spatial and textual learning in supervised finetuning (SFT) and reinforcement learning (RL) by considering two additional tasks during the training beyond the chart question answering (CQA): (i) chart element grounding, which improves the model's ability to localize and identify specific visual components, explicitly learning the spatial structure of the chart and (ii) chart-to-code generation, which enhances the model's understanding of the underlying textual details of the chart, also implicitly learning the chart layout from Python code. This spatial and textual learning approach ensures that models develop both the textual analytical capabilities needed for data interpretation and the spatial reasoning required for chart elements or structures identification.

% 6. what we proposed (the dataset)
To facilitate spatial and textual learning, we propose the START-Dataset with a novel data-generation pipeline that bypasses the template-driven visual simplicity, limited layout/style diversity, and chart-type distribution mismatch characteristic of code-based synthetic datasets~\cite{yang2025effective, he2024distill}. Our method employs an MLLM to translate real chart images~\cite{li2024multimodal} into executable chart Python code, preserving the visual complexity and diversity of real-world charts and reconstructing the underlying data representation of the chart. Based on the recovered chart code, we propose to find the location of the chart element by evolving the Python code with a large language model (LLM), addressing the issue that the existing MLLM~\cite{bai2025qwen2}, grounding~\cite{liu2024grounding}, or segmentation model~\cite{ravisam} could not ground the chart element correctly. 
% and introducing the first chart element grounding data source, facilitating the chart element grounding task.
These precise chart elements' locations, coupled with the Python code for rendering the chart image, enable the spatial and textual learning for the chart understanding.
To evaluate the model's understanding of the chart visual structure, we propose the Chart-Spatial understanding Benchmark (CS-Bench) to fill in the missing piece in comprehensive chart understanding.

Our comprehensive evaluation demonstrates that START achieves substantial improvements over the base model across multiple chart understanding benchmarks in both reinforcement learning (RL) and supervised fine-tuning (SFT) paradigms. Specifically, START-RL-7B outperforms the previous best~\cite{chen2025chart} by 42.7, 14.7, 1.5, 2.1, and 35.7 on ChartMimic~\cite{yangchartmimic}, CharXiv-descriptive~\cite{wang2024charxiv}, CharXiv-reasoning,  ChartQAPro~\cite{masry2025chartqapro}, and CS-Bench, respectively.

\medskip
Our main \textbf{contributions} are summarized as follows.
\begin{itemize}
    \item We propose START, a spatial and textual learning framework for chart understanding in SFT and RL, leveraging code learning and chart element grounding to enhance MLLM's understanding toward the chart.
    \item To facilitate the spatial-textual training, we create the START-Dataset and a new dataset construction pipeline by converting the real chart images to Python codes, then evolving the code with LLM to obtain chart element locations.
    We propose the Chart Spatial understanding Benchmark (CS-Bench) to evaluate the MLLM's spatial understanding of the chart.
    \item START achieves substantial improvement across different model scales and various benchmarks and outperforms the previous best by a clear margin.
\end{itemize}

\section{Related Works}
\label{sec:related_works}

\noindent \textbf{Multi-modal Large Language Model.}
%The general MLLM.
Instruction tuning with visual and text data~\cite{liu2023visualinstructiontuning, liu2023improvedllava, liu2024llavanext} has led to a surge of interest in using the Large Language Model for Visual Understanding. 
% Many of these models share a common design, where visual features are extracted using a pre-trained visual encoder, projected into the text latent space of an LLM, and subsequently processed by the pre-trained LLM to generate responses. 
Later works, LLaVA-OneVision~\cite{li2024llava}, Qwen2.5-VL~\cite{bai2025qwen2}, scaling up the training with visual input from different modalities (images, videos, etc), substantially improve the model performance on different visual understanding tasks and introduce the concept of Multi-modal Large Language Model (MLLM)~\cite{chen2024expanding, ye2024mplugowl3longimagesequenceunderstanding}.
%The reasoning MLLM. 
Recent success of DeepSeek-R1~\cite{guo2025deepseek} shows that reason-before-answering coupled with reinforcement-learning~\cite{shao2024deepseekmath} yields substantial gains on difficult math~\cite{lightman2023lets, aime_1983_2024} and programming benchmarks~\cite{penedo2025codeforces, jain2024livecodebench}. Motivated by it, vision-reasoning models such as Vision-R1~\cite{huang2025vision} and R1-OneVision~\cite{yang2025r1} transplant these think-then-answer strategies to the multimodal setting, improving MLLM performance on challenging visual reasoning benchmarks~\cite{lumathvista, zhang2024mathverse}.
% Later, the success of o3 and o4-mini ~\cite{o3openai} reveals that the interaction (cropping, zooming-in, flipping) with the visual input during the reasoning can further enhance the model's visual understanding. Thus, encouraging the agentic pipeline to be used during visual reasoning.

In our work, we propose spatial and textual learning for both supervised finetuning and reinforcement learning, enabling comprehensive chart learning for MLLM on different training paradigms.

%The chart-specific MLLM.
\noindent \textbf{Chart Understanding.}
Early research in chart understanding~\cite{kafle2018dvqa, chaudhry2020leaf, methani2020plotqa, luo2021chartocr, kantharaj-etal-2022-chart, poco2017reverse, cliche2017scatteract} primarily explored structural designs, combining Convolutional Neural Network (CNN) encoders~\cite{he2016deep} with Recurrent Neural Network (RNN) decoders~\cite{rumelhart1985learning, hochreiter1997long}. Subsequent works~\cite{masry2022chartqa, singh-shekhar-2020-stl} introduced multi-stage pipelines and transformer-based architectures, yielding stronger performance. More recently, multimodal large language models (MLLMs) have become the dominant paradigm for chart understanding. Within this line, some studies enhanced instruction-tuning data~\cite{han2023chartllama, masry-etal-2024-chartinstruct, masry2024chartgemma, xuchartmoe}, while others improved MLLM reasoning via chart-to-text tasks~\cite{zhang-etal-2024-tinychart, meng2024chartassistant} or reinforcement learning~\cite{chen2025chart, masry2025bigcharts}. A parallel thread of research has focused on chart-to-code generation~\cite{jia2025chartreasoner, chen2025breaking, tan2025chartmaster}.

From a data perspective, early works on chart image synthesis~\cite{kahou2017figureqa, kafle2018dvqa, methani2020plotqa} relied on parameterized templates to produce synthetic variants. To expand both scale and diversity, later methods~\cite{he2024distill, yang2025effective, han2023chartllama} leveraged LLMs to evolve Python code and render new charts. In parallel, real-world chart images~\cite{masry2022chartqa, masry-etal-2023-unichart, masry-etal-2024-chartinstruct, li2024multimodal} were collected from the web to better capture natural chart distributions. Complementary to data construction, recent efforts introduced comprehensive benchmarks for chart question answering~\cite{masry2022chartqa, masry2025chartqapro, wang2024charxiv, liu2023mmc, kafle2018dvqa, kahou2017figureqa, vogel2025refchartqa} and chart-to-code/text generation~\cite{yangchartmimic, wu-etal-2025-plot2code, xia2024chartx}.

However, prior research largely overlooks the importance of spatial structure in chart understanding. Motivated by the chart's property, we introduce spatial and textual learning for chart understanding, which explicitly models the spatial structures of charts. To enable this, we propose a new dataset construction pipeline that lies between synthetic and real chart approaches: it generates diverse synthetic charts while also providing the corresponding Python codes. Furthermore, we design a new benchmark for chart spatial understanding, filling in the gap in comprehensive chart evaluation.

\noindent \textbf{Spatial and textual understanding.}
Previous works show that both spatial and textual learning enhance visual understanding. MDETR~\cite{kamath2021mdetr} demonstrates that improving object localization boosts performance on vision-language tasks. V*~\cite{wu2024v} and OpenAI o3~\cite{o3openai} further reveal that accurate object localization and zooming improve fine-grained visual understanding. Similarly, spatial reasoning has been shown to benefit downstream tasks such as navigation~\cite{roh2022languagerefer}.
Others research highlights the role of textual learning, where visual scenes are converted into text for improved reasoning. Examples include document-to-text~\cite{singh2021textocr, singh2019towards}, video captioning for video understanding~\cite{wang2024vamos, zhang2025silvr}, and chart-to-text conversion for chart reasoning~\cite{jia2025chartreasoner}.

Inspired by the dual spatial–textual nature of charts, our work integrates both perspectives for the first time and demonstrates that spatial and textual understanding significantly improves MLLMs on chart understanding.

\begin{figure}[t!]
    \centering
    \includegraphics[width=1.0\linewidth]{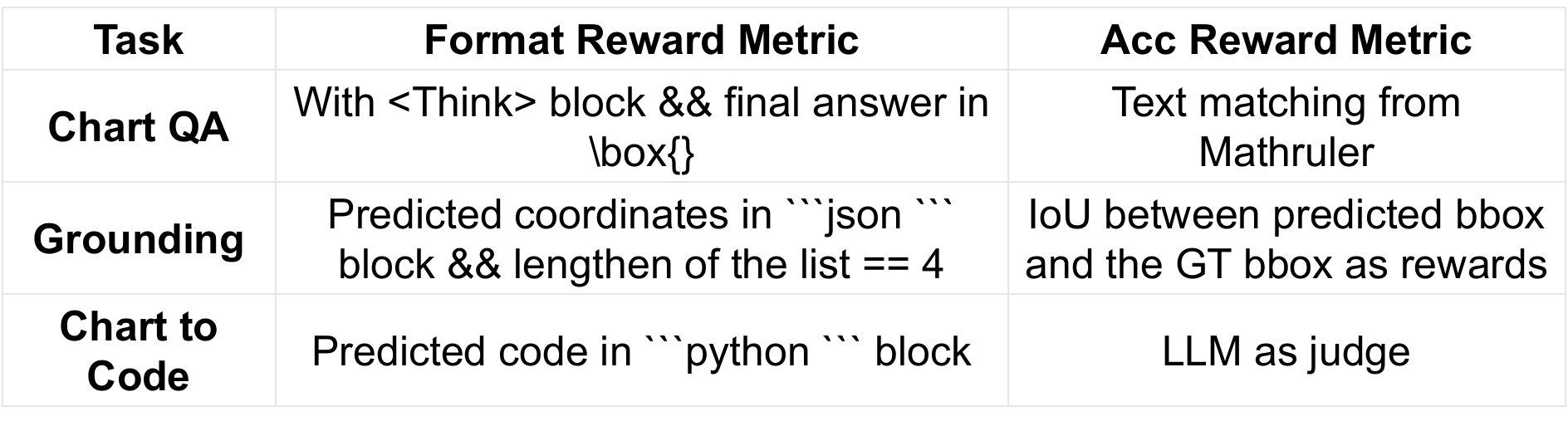}
    \vspace{-2em}
    \caption{START's reward design in reinforcement learning.}
    \label{fig:learning_overview_and_rewards}
    \vspace{-2em}
\end{figure}

\begin{figure*}[t!]
    \centering
    \includegraphics[width=0.95\linewidth]{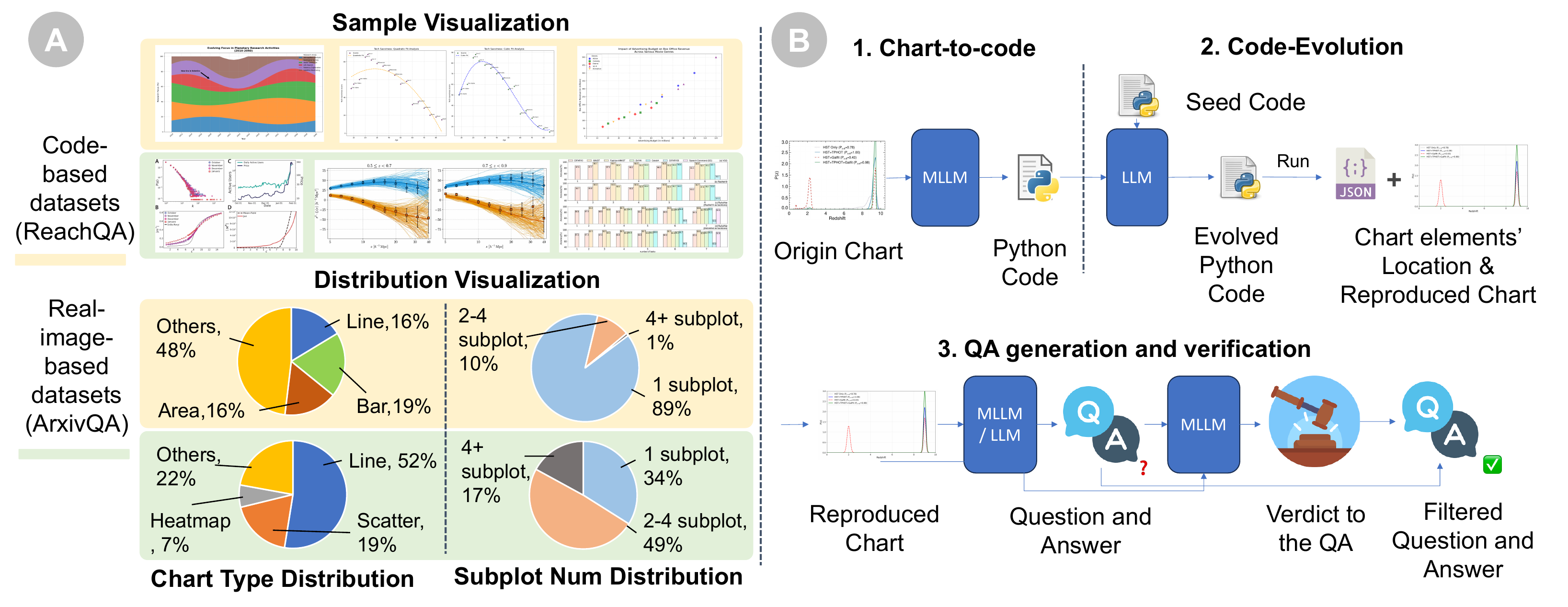}
    \vspace{-1em}
    \caption{A: The analysis of the existing chart datasets and B: the overview of the START-dataset generation pipeline.}
    \label{fig:dataset_motivation_and_pipeline}
    \vspace{-1.5em}
\end{figure*}

\section{Spatial and Textual Learning for Chart Understanding (START)}\label{sec:dataset}
In this section, we first introduce the definition of spatial and textual learning for chart understanding in section~\ref{chart_understanding}. To facilitate the spatial and textual learning, we construct the START-Dataset and introduce the details in section~\ref{start_dataset}. Finally, we introduce the Chart Spatial understanding Benchmark (CS-Bench) in section~\ref{cs_Benchmark}

\subsection{Method}~\label{chart_understanding}
Motivated by the dual property of the chart, which has a structured spatial layout and the corresponding textual data representation, we propose spatial and textual learning for chart understanding with a multi-modal large language model (MLLM). 

\medskip
\noindent\textbf{Background of MLLM.} 
Multi-modal Large Language Models (MLLMs) extend the capability of Large Language Models (LLMs) by integrating visual encoders with text-based transformers. The visual encoder maps raw pixels into a latent feature space, which is then aligned with textual embeddings in the shared language model backbone. This alignment enables the model to process and reason over both visual and textual inputs jointly.  

Formally, given an image $I$ and a text prompt $q$, visual encoder and the adapter $E_v$ and text encoder $E_t$ encodes them into vision embeddings $z_I$ and text embedding $z_q$:
\begin{equation}
z_I = E_v(I), \quad z_q = E_t(q),
\label{eq:mllm}
\end{equation}
These embeddings are then fused in a shared feature space and decoded by the LLM backbone $D$ to generate outputs:
\begin{equation}
y = D(z_I, z_q).
\end{equation}
Here, $y$ can represent different outputs in text, such as localized grounding coordinates, natural language responses, or programmatic code, depending on the task instruction. This unified framework allows MLLMs to support spatial-textual reasoning in chart understanding.

\medskip
\noindent\textbf{Spatial Learning} for visual inputs~\cite{liu2024grounding, kamath2021mdetr} aims to identify object instances relevant to a given question from the input image, capturing the relationships among these objects, and thereby improving scene understanding.
In the context of chart understanding, we treat each chart element (e.g., title, legend, subplot, etc) as an instance. Spatial learning enables the model to better understand the visual structure and layout of the chart and to establish a mapping between chart concepts and their corresponding locations.
Formally in MLLM's chart understanding, given an image $I$ and a question $q_s$ related to the chart’s spatial structure, MLLM predicts:
\begin{equation}
z_{q_s} = E_t(q_s); y_s = D(z_I,z_{q_s})
\end{equation}
where $y_s$ denotes the location of the answer within the chart, and $z_I$ is defined in equation~\ref{eq:mllm}.

\medskip
\noindent\textbf{Textual Learning} for visual inputs~\cite{jia2025chartreasoner, singh2019towards} focuses on bridging the details in visual inputs with their textual counterparts, thereby recovering the underlying data semantics encoded in the visual signal.
For chart understanding, we consider the Python code used to render a chart as the textual representation of its underlying data and structure. Textual learning builds connections between chart element locations, their associated values, and the corresponding formulas or rendering instructions in the code.
Formally in MLLM's chart understanding, given an image $I$ and a prompt $q_t$ that instructs the model to generate its textual representation for the input chart image, MLLM predicts:
\begin{equation}
z_{q_t} = E_t(q_t); y_t = M(z_I,z_{q_t})
\end{equation}
where $y_t$ is the textual representation (i.e., Python code) of the input chart image, and $z_I$ shares the same definition as the one in equation~\ref{eq:mllm}.
By incorporating textual learning, the model gains the ability to capture fine-grained details from the chart image, thereby enhancing its overall understanding of the chart.

\medskip
\noindent\textbf{Learning Designs.} We apply spatial and textual learning in two learning paradigms, supervised finetuning (SFT) and reinforcement learning (RL).
We train MLLMs with spatial and textual learning, along with the chart question answering (CQA). This ensures that the MLLM leverages spatial and textual learning to enhance chart understanding while preserving its original conversational capabilities. For SFT, we update the model by minimizing the negative log likelihood loss. In RL, consider updating the model with Group Relative Policy Optimization~\cite{shao2024deepseekmath} (GRPO).

For the reward design in the RL training,  the reward consists of two parts: 1. the accuracy reward $R^{acc}$ and 2. the format reward $R^{format}$.
For the accuracy reward, we apply a different metric to judge the reward for the prediction.
For CQA samples, we use the string matching from the Mathruler~\cite{mathruler} to judge the correctness of the prediction. We use the IoU value between the predicted bounding box and the ground truth bounding box as the reward for the spatial learning samples. 
We apply an LLM as a judge to grade the generated Python code for the textual learning samples. 
For the format reward, we mainly apply text matching with a regular expression on model predictions. The details can be found in Figure~\ref{fig:learning_overview_and_rewards}.
The final reward $R_i$ for the prediction $o_i$ is defined as: 
\begin{equation}
R_i = a \times R^{acc}_i + (1-a) \times R^{format}_i
\end{equation}
We use $a=0.9$ in our case. Please refer to the supplementary material for more information on the learning design.

\begin{figure*}[t!]
    \centering
    \includegraphics[width=1.0\linewidth]{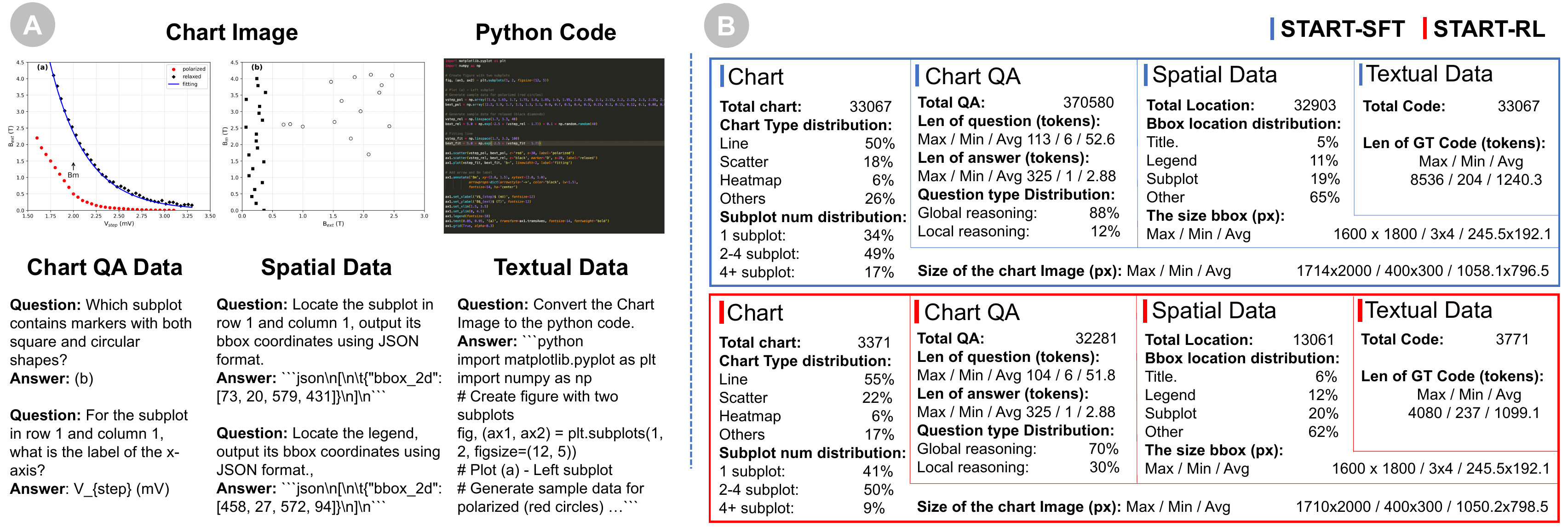}
    \vspace{-2em}
    \caption{A: The dataset sample visualization and B: the START-SFT and START-RL dataset statistics.}
    \label{fig:dataset_stat}
    \vspace{-1.5em}
\end{figure*}

\subsection{START-Dataset}~\label{start_dataset}
To enable effective spatial and textual training, we need a dataset comprising high-quality chart images $I$ paired with (1) $D_s$ capturing the chart’s visual layout (i.e., chart element locations), which will be used in the spatial learning. (2) its underlying data representation $D_t$ (i.e., the Python code used to render the chart image), which facilitates the textual learning. (3) instruction learning question-answer pairs $D_q$, which will be used in the chart question answering learning.

\medskip
\noindent\textbf{Motivation and Chart Dataset Analysis.} 
Existing chart datasets generally fall into two categories: (1) synthesis-image-based and (2) real-image-based. Synthesis-image-based datasets, such as ReachQA~\cite{he2024distill}, are generated using Large Language Models (LLMs). Starting from a set of seed chart codes, LLMs are prompted to combine or modify these seeds to produce diverse variants with greater visual complexity. In contrast, real-image-based datasets, such as ArxivQA~\cite{li2024multimodal}, in which charts are collected online, reflects the distribution of charts in real-world scenarios.

Figure~\ref{fig:dataset_motivation_and_pipeline}-A presents samples from both ReachQA and ArxivQA, along with their chart-type and subplot-count distributions. Visually, charts from code-based datasets tend to be simpler, with fewer subplots and less detail per subplot (e.g., fewer lines or bars) compared to their real-image counterparts. Statistically, code-based datasets also differ in chart-type distribution and overwhelmingly feature single-subplot charts—further highlighting the domain gap between the two dataset types.

However, real-image-based datasets cannot directly serve our needs due to their lack of underlying data representations—such as data tables or the Python code required to render the charts. This limitation motivates us to develop our own dataset that preserves the visual complexity of real-world charts and provides their underlying data representations.
As shown in Figure~\ref{fig:dataset_motivation_and_pipeline}-B, our dataset construction pipeline consists of three steps: 1. to prepare the spatial learning data $D_t$, we leverage the chart to code conversion to obtain Python code along with the reconstructed chart. 2. code evolution for obtaining the chart elements' location and spatial learning data $D_s$. 3. question-answer pairs generation and verification, preparing the data for chart question answering $D_q$.

\medskip
\noindent\textbf{Textual Learning Data Construction.} For textual learning, we prepare the (chart image)-(chart code) pairs as the textual learning data.
To obtain chart images that share a similar distribution to the real chart data, and also come with the Python code, we converted the chart images from the ArxivQA~\cite{li2024multimodal} to Python code and then rendered the reproduced chart image.
To convert the chart image to Python code, we explored different approaches.
Our exploration shows that using strong MLLM to directly convert the chart images to Python codes yields the most authentic reproduced chart images within a reasonable budget. After we obtain the codes, we then convert the codes to question-answer pairs with fixed templates. This obtains the training data $D_c$ for the textual learning. 
Please see more details in the supplementary material A.1.

\medskip
\noindent\textbf{Spatial Learning Data Construction.} For spatial learning, we need to obtain the chart element locations on the image as spatial learning data.
Some existing works like ChartQA~\cite{masry2022chartqa}, FigureQA~\cite{kahou2017figureqa} leverage SVG parsing or a fixed template to find the chart element location, but these strategies become infeasible when the chart image contains multiple subplots, diverse layouts, and densely packed details. Another idea for finding the chart element locations is to use a multi-modal large language model (MLLM)~\cite{bai2025qwen2}, a grounding model~\cite{liu2024grounding}, or a segmentation model~\cite{ravisam}. However, these models could not ground the chart element correctly, given that they barely trained on the chart images, which makes them hard to understand the chart concept. 

To ensure the chart element's location is consistent with the chart image, we generate the chart element locations while rendering the chart image. We propose to evolve the Python code with a Large Language Model (LLM) to obtain the location of the chart elements.
Specifically, for each chart type, we prepare seed codes that leverage the matplotlib built-in function to obtain the chart element's location on the rendered chart image, and further save the obtained chart element location into a JSON file. 
We then use these seed codes as examples to prompt the LLM to evolve the Python code generated in the previous section. 

Running the evolved code will generate a rendered chart image and a JSON file that stores all locations of the chart element. We then convert the chart element locations to a question-answer pairs with fixed templates, obtaining the data $D_g$ that will be used in the spatial learning. Please see more details in the supplementary material A.2.
% The visualization of the chart element locations is shown in figure~\ref{fig:visualization_tsne_and_grounding}-B.

\begin{figure*}[t!]
    \centering
    \includegraphics[width=1.0\linewidth]{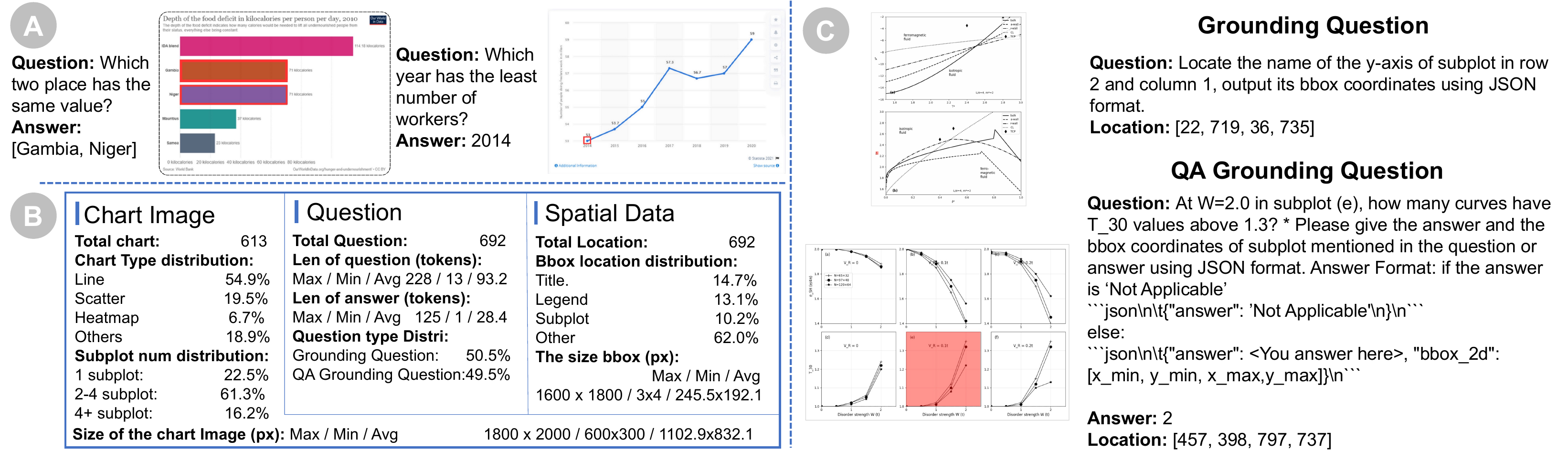}
    \vspace{-2em}
    \caption{A: The samples from refChartQA~\cite{vogel2025refchartqa}, the locations are related to limited types of chart components, and focus on single-subplot chart images. B: the CS-Bench statistics. C: the samples from CS-Bench with the visualized target region under a red mask.}
    \label{fig:benchmark_all}
    \vspace{-1.5em}
\end{figure*}

\medskip
\noindent\textbf{QA Learning Data Construction.} For chart question answer (CQA) learning, we generate high-quality question-answer pairs related to a chart image.
We prompt the MLLM to generate questions that do not require external domain knowledge but instead rely solely on the chart content. This prevents the model from simply memorizing domain-specific answers and encourages the model to learn reasoning over general chart structures and content.

To ensure the QA quality, we prompt a strong MLLM to detect the hallucinated questions that refer to non-existent elements or those beyond the MLLM’s capacity (e.g., precisely counting 200 dots) and verify the correctness of each answer. Based on these verdicts, we filter the QA pairs to obtain the final dataset for the CQA learning $D_q$.
We further categorize the questions into global reasoning or local reasoning, based on whether they need reasoning over multiple chart elements. Please see more details in the supplementary material A.3.

\medskip
\noindent\textbf{Dataset Visualization and Statistics.} \label{dataset_stat_and_distri}
We show the sample visualization in Figure~\ref{fig:dataset_stat}-A. Each chart image is paired with the Python code, which is used to render it, the chart question answering data, the spatial learning data, and the textual learning data.
We combine the data from chart question answering, spatial learning, and textual learning as the Supervised-Finetuning Dataset (START-Dataset-SFT), and we sample the Reinforcement Learning Dataset (START-Dataset-RL) from the START-Dataset-SFT based on the question difficulties.
Figure~\ref{fig:dataset_stat}-B shows the statistics of START-SFT and START-RL.

\subsection{CS-Bench}\label{cs_Benchmark} 

\noindent\textbf{Motivation.}
Existing chart benchmarks primarily focus on chart understanding~\cite{masry2022chartqa, wang2024charxiv, masry2025chartqapro} or chart-to-text generation~\cite{yangchartmimic, xia2024chartx}, while the evaluation of a model’s understanding to chart's spatial structure remains underexplored. RefChartQA~\cite{vogel2025refchartqa} takes a step in this direction by requiring models to ground chart elements relevant to a question. However, its grounding is limited to simple cases—such as locating points on a line or bars in a bar chart—where bounding boxes are strongly influenced by human annotation bias and lack explicit chart-related semantics. Moreover, it is restricted to single-subplot charts (see Figure~\ref{fig:benchmark_all}-A for examples).
To address these limitations, we introduce the Chart Spatial understanding Benchmark (CS-Bench), which specifically evaluates MLLMs on their ability to reason about and localize spatial structures in charts, thereby filling a crucial gap in the comprehensive assessment of chart understanding.

\medskip
\noindent\textbf{Data Construction Pipeline.} To ensure the quality of our benchmark, CS-Bench is constructed from a held-out set of chart images rendered by the evolved code from our START-Dataset pipeline. This strategy guarantees that the charts in the benchmark reflect the complexity and distribution of those found in real-world scenarios.

CS-Bench incorporates two question types to rigorously assess a model's spatial reasoning capabilities:
1) Grounding Questions: These directly prompt a model to localize specific chart elements, such as a subplot, an axis tick value, or the title. By converting known element locations into question-answer pairs using fixed templates, these questions provide a direct evaluation of the model's core spatial understanding.
2) QA Grounding Questions: These present a more complex task where the model must first answer a question related to the chart's content and then localize the visual elements that are mentioned in the question or referred to in the answer. This evaluates a deeper level of comprehension, specifically the model's ability to connect concepts within the question or answer to the correct chart structures.
To maintain the highest standard of quality and reliability, all question-answer pairs and their corresponding element locations in the benchmark have been manually verified.
We present the benchmark statistics the Figure~\ref{fig:benchmark_all}-B and visualize the samples in the benchmark in Figure~\ref{fig:benchmark_all}-C.
The image distribution in CS-Bench is similar to real-world chart collections, which are dominated by images containing multiple subplots. Localizing chart elements in these complex layouts is a significant challenge, thereby demonstrating the benchmark's ability to rigorously evaluate a model's performance.

\medskip
\noindent\textbf{Metrics.}
Following the existing spatial understanding benchmarks (e.g. image grounding~\cite{lin2014microsoft, young2014image}, video grounding~\cite{krishna2017dense, gao2017tall, zhou2018towards}),
we report the recall of ground-truth (GT) bounding boxes at an Intersection over Union (IoU) threshold of 0.3, evaluated on both grounding and QA grounding questions. As an auxiliary metric, we also report the accuracy on QA grounding questions.
We provide more details in the supplementary material section B.

\begin{table*}[t]  
\centering  
\scalebox{0.8}{
\begin{tabular}{lc|cc|c|c|c|cc}  
\toprule
\multirow{2}{*}{Method}  & \multirow{2}{*}{Model size} & \multicolumn{2}{c|}{CharXiv~\cite{wang2024charxiv}} & \multirow{2}{*}{ChartQA~\cite{liu2023mmc}}  & \multirow{2}{*}{ChartQAPro~\cite{masry2025chartqapro}} & \multirow{2}{*}{ChartMimic~\cite{yangchartmimic}} &  \multicolumn{2}{c}{CS-Bench}\\
                         &                             & desc     & rea                                      &     &   & & R@0.3 & acc \\ 
    \midrule
    \multicolumn{6}{l}{\it \small\textbf{General MLLMs}} \\
     Qwen2.5-VL~\cite{bai2025qwen2}  & 3B &  59.7 \textcolor{gray}{(58.6)} & 32.1  \textcolor{gray}{(31.3)} & 84.0 & 32.9 &  29.3 & 16.2 & 45.0 \\
     Qwen2.5-VL~\cite{bai2025qwen2}  & 7B & 66.6 \textcolor{gray}{(73.9)} & 41.4 \textcolor{gray}{(42.5)} & 87.3 & 41.3  & 40.2 & 19.3 & 50.2 \\
    \multicolumn{6}{l}{\it \small\textbf{Chart-specific MLLMs}} \\
     TinyChart~\cite{zhang-etal-2024-tinychart}  & 3B & - & 8.3 & 83.6 & 13.2  & - & - & -\\
     ChartGemma~\cite{masry2024chartgemma}       & 3B & - & 12.5 & 80.1 & 6.8  & - & - & -\\
     ChartReasoner~\cite{jia2025chartreasoner}   & 7B & - & - & 86.3 & 39.9  & - & - & -\\
     ECD~\cite{yang2025effective}                & 7B & 74.2 & 40.2 & 85.3 & 22.5 & 46.8 & 3.6 & 7.3\\
     Chart-R1~\cite{chen2025chart}               & 7B & 62.0 & 45.2 \textcolor{gray}{(46.2)} & \textbf{91.0} & 44.0  & 18.7 & 9.6 & 54.5 \\
    % \multicolumn{6}{l}{\it \small\textbf{Naive SFT and RL}} \\
    %     SFT                                     & 3B & 61.1 & 35.6 & 84.1 & 32.8 & 30.9 & 20.6 & 49.0 \\ 
    %     RL                                      & 3B & 68.3 & 37.0   & 84.7 & 36.4 & 32.0 & 22.6 & 57.8 \\
    %     SFT                                     & 7B & 66.7 & 43.3 & 87.5 & 41.6 & 41.9 & 15.9 & 52.3 \\
    %     RL                                      & 7B & 70.1 &  44.8 & 84.8 & 41.8 & 55.3 & 37.0 & 60.2 \\
    \midrule
    \multicolumn{6}{l}{\it \small\textbf{Ours models}} \\
     START-SFT                          & 3B & 63.1 & 35.9 & 84.4 & 34.2 & 31.5 & 26.9 & 58.8 \\
     START-RL                           & 3B & 72.2 & 40.0 & 84.8 & 38.2 & 45.3 & 41.3 & 60.5  \\
     START-SFT                          & 7B & 66.9 & 44.0 & 87.6 & 41.8 & 50.7 & 31.0 & 57.6 \\
     START-RL                           & 7B & \textbf{77.6} & \textbf{46.7}  & 88.8 & \textbf{46.3} & \textbf{63.8} & \textbf{45.3} & \textbf{62.3} \\
    
     \bottomrule

\end{tabular}
}
\vspace{-1.5mm}
\caption{Results of general chart understanding. We follow~\cite{yang2025effective} to present the reproduced number of Qwen models and add the number from Qwen technical report~\cite{bai2025qwen2} in \textcolor{gray}{gray} as reference, similarly for Chart-R1. START shows substantial improvement over the base models and outperforms the previous best, Chart-R1~\cite{chen2025chart}, on CharXiv, ChartQAPro, ChartMimic and CS-Bench by a clear margin.}
\vspace{-3mm}
\label{tab:chart_understanding}  
\end{table*}

\begin{table}[t]  
\centering  
\scalebox{0.75}{
\begin{tabular}{l|cc|c|c|cc}  
\toprule
\multirow{2}{*}{Method}  & \multicolumn{2}{c|}{CharXiv~\cite{wang2024charxiv}} & \multirow{2}{*}{CQAPro~\cite{masry2025chartqapro}} & \multirow{2}{*}{CMimic~\cite{yangchartmimic}} & \multicolumn{2}{c}{CS-Bench}  \\
                         & desc     & rea                                      &     &  & R@0.3 & acc  \\ 
    \midrule
    \multicolumn{6}{l}{\it \small\textbf{SFT}} \\
     Q                           & 61.1 & 35.6 & 32.8 & 30.9 & 20.6 & 49.0 \\
     Q+C                         & 60.5 & 35.5 & 34.1 & 32.1 & 22.2 & 51.1 \\
     Q+C+G                       & 63.1 & 35.9 & 34.2 & 31.5 & 26.9 & 58.8 \\
     \multicolumn{6}{l}{\it \small\textbf{RL}} \\
     Q                           & 68.3 & 37.0   & 36.4 & 32.0 & 22.6 & 57.8 \\
     Q+C                         & 72.3 & 38.7 &  37.5 & 43.0 & 22.8 & 58.6 \\
     Q+C+G                       & 72.2 & 40.0 & 38.2 & 45.3 & 41.3 & 60.5 \\
     \bottomrule
\end{tabular}
}

\vspace{-1mm}
\caption{Ablation Study of chart question answering (Q), chart-to-code (C), and chart element grounding (G) on CharXiv, ChartQAPro (CQAPro), ChartMimic (CMimic), and CS-Bench. Adding chart-to-code enhances the textual understanding of the chart, obtaining consistent improvement on CQAPro and CMimic, while adding chart element grounding improves the spatial understanding of the chart, boosting the model's performance on different benchmarks.}
\vspace{-4mm}
\label{tab:ablation}  
\end{table}

% \begin{figure*}[t!]
%     \centering
%     \includegraphics[width=0.85\linewidth]{figures/visualization_tsne_and_grounding.pdf}
%     \vspace{-1em}
%     \caption{The visualization data distribution, data sample, and the different chart element grounding methods. A: shows t-SNE map of the chart image feature from our dataset and other datasets. 
%     B: visualizes the bounding boxes in our chart element grounding dataset.}
%     \label{fig:visualization_tsne_and_grounding}
%     \vspace{-1.5em}
% \end{figure*}

\begin{figure*}[t!]
    \centering
    \includegraphics[width=0.9\linewidth]{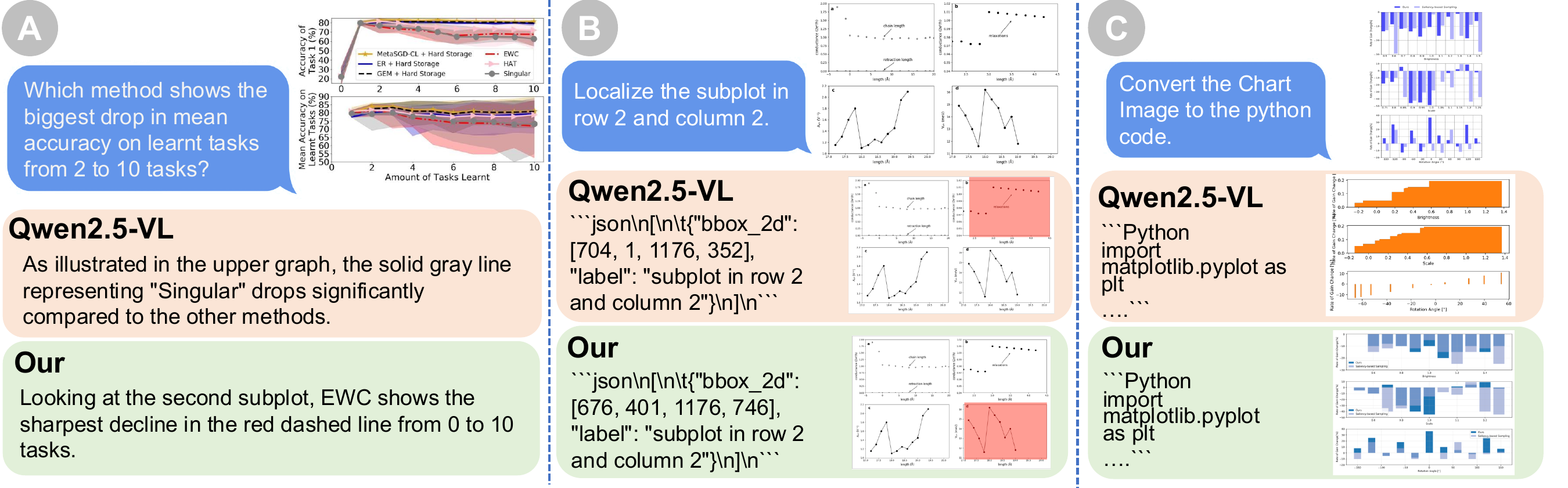}
    \vspace{-1em}
    \caption{The visualization of the predictions from START verse Qwen2.5-VL~\cite{bai2025qwen2}. Benefit from the spatial and temporal learning, START produces better predictions in chart question answering (Subplot A), chart element grounding (Subplot B), and chart-to-code (Subplot C), reflecting the enhancement in MLLM's spatial and textual understanding toward the charts.}
    \label{fig:qa_code_grounding}
    \vspace{-1.5em}
\end{figure*}

\section{Experiments and Results}\label{sec:experiments}
In this section, we present the evaluation of the START's performance across various chart understanding tasks, as described in Section~\ref{Chart_understanding_benchmark}. We then conduct an ablation study in Section~\ref{ablation_study} to demonstrate the effectiveness of chart element grounding and chart-to-code components. Finally, in Section~\ref{analysis_and_visualization}, we provide visualizations and offer a detailed analysis of our method.

\subsection{Results on Chart Understanding} \label{Chart_understanding_benchmark}

\noindent\textbf{Benchmarks.}
To comprehensively evaluate START, we use CharXiv~\cite{wang2024charxiv}, ChartQA~\cite{masry2022chartqa}, ChartQAPro~\cite{masry2025chartqapro}, ChartMimic~\cite{yangchartmimic} and CS-Bench proposed by us as benchmarks. 
The CharXiv-reasoning~\cite{wang2024charxiv} and ChartQAPro~\cite{masry2025chartqapro} test multi-step reasoning, while the CharXiv-descriptive and ChartQA~\cite{masry2022chartqa} splits focus on simple perception and reasoning. ChartMimic~\cite{yangchartmimic} evaluates chart-to-code translation to assess understanding of the underlying data and visual details. 
Our CS-Bench evaluates MLLM's chart spatial understanding. We report scores for CharXiv and ChartMimic, accuracy (acc) for ChartQA and ChartQAPro, and recall@0.3 and acc for CS-Bench.

\noindent\textbf{Implementation Details.}
We initiate the Supervised Finetuning (SFT) training with Qwen2.5-VL-3B and 7B model~\cite{bai2025qwen2} checkpoints, given that they have promising reasoning ability that could be elicited during the RL training~\cite{huang2025vision, yang2025r1, feng2025video} and reasonable grounding performance, thanks to its high-quality pretraining. We train the models with SFT on the START-SFT dataset for 1 epoch. We then initialize the Reinforcement Learning (RL) with the SFT checkpoint, and train the model with RL with the START-RL dataset for 100 steps. For the SFT, we train the model with learning 1e-6 with a 0.1 warm-up ratio and cosine learning rate decay. We set the global batch size to 128. For the RL training, we train the model with a learning rate of 1e-6, rollout batch size 512, and rollout number of 5. During the training, we set the micro-batch size per device to 4.

\noindent\textbf{Baselines.}
We mainly consider 2 different types of baselines: 1. the MLLM that targets general image understanding: for instance, Qwen2.5-VL-3B and 7B model~\cite{bai2025qwen2} 
%and intern-VL2.5-4B and 8B ~\cite{chen2024expanding}, 
2. Chart-Specific MLLM, for instance, TinyChart~\cite{zhang-etal-2024-tinychart}, ChartGemma~\cite{masry2024chartgemma}, ChartReasoner~\cite{jia2025chartreasoner}, ECD~\cite{yang2025effective} and Chart-R1~\cite{chen2025chart}.
% 3. Naive SFT and RL training with our Chart QA dataset as a baseline to demonstrate the effectiveness of the spatial-textual learning. 
Noticed that there is a gap between the reproduced and the official number for the Qwen models on CharXiv, we follow~\cite{yang2025effective} to present the reproduced number and provide the official number~\cite{bai2025qwen2} in \textcolor{gray}{gray}.

\noindent\textbf{Results.}
We present the results in Table~\ref{tab:chart_understanding}. START obtains consistent improvement over the base model on both SFT and RL settings. 
Compared with the Qwen base models, START-RL-3B and START-RL-7B show (12.5/10.1), (7.9/5.3), (5.3/4.8), (16/21.2), (25.1/26) on CharXiv descriptive, CharXiv reasoning, ChartQAPro, ChartMimic, and CS-Bench, respectively.
START-RL-7B surpasses the previous best Chart-R1-7B by 14.7, 1.5, 2.1, 42.7 and 35.7 on CharXiv descriptive, CharXiv reasoning, ChartQAPro, ChartMimic and CS-bench, respectively, while lagging behind on ChartQA, as we did not use ChartQA during training and instead demonstrated zero-shot performance.
% START also shows consistent improvement over the Naive SFT and RL baseline, which demonstrates effectiveness of the spatial-textual learning.

\subsection{The Ablation Study.} \label{ablation_study}
We conduct an ablation study for learning designs in spatial-textual learning. In this study, we use CharXiv, ChartQAPro, ChartMimic, and CS-Bench as benchmarks to demonstrate the model's performance. We use the Qwen2.5-VL 3B model to experiment with the same training details mentioned in section~\ref{Chart_understanding_benchmark}. 

\medskip
\noindent\textbf{Effectiveness of the Spatial and Textual Tasks.}
We train the model with different task combinations: 1. Chart question answer (CQA) only (Q), CQA + Chart-to-Code (Q+C), and CQA + Chart-to-Code + Chart element grounding (Q+C+G) on SFT and RL, respectively. The experiment results in Table~\ref{tab:ablation} show that adding the Chart-to-Code task improves the textual understanding of the model and obtains consistent improvement on ChartQAPro and ChartMimic, which requires capturing the details on the chart images. Further adding the chart element grounding task improves the model's spatial understanding of the chart and obtains substantial improvement on CharXiv and CS-Bench, in which questions require localizing the chart element correctly.
Surprisingly, when adding grounding in the RL setting, we see further improvement in ChartQAPro and ChartMimic, which demonstrate that spatial and textual learning complement each other.

\begin{table}[t]  
\centering  
\scalebox{0.7}{
\begin{tabular}{l|cc|c|c|cc}  
\toprule
\multirow{2}{*}{Thinking}  & \multicolumn{2}{c|}{CharXiv~\cite{wang2024charxiv}} & \multirow{2}{*}{CQAPro~\cite{masry2025chartqapro}} & \multirow{2}{*}{CMimic~\cite{yangchartmimic}} & \multicolumn{2}{c}{CS-Bench}  \\
                         & desc     & rea                                      &     &  & R@0.3 & acc  \\ 
    \midrule
     on QA only                         & 69.7 & 38.7 & 36.7 & 40.2 & 39.6 & 58.0 \\
     on all tasks                       & 72.2 & 40.0 & 38.2 & 45.3 & 41.3 & 60.5 \\
     \bottomrule
\end{tabular}
}
\vspace{-2mm}
\caption{Ablation study for applying think-before-answer in training. Applying thinking in all tasks yields better results.}
\vspace{-4mm}
\label{tab:ablation_on_thinking}  
\end{table}

\medskip
\noindent\textbf{Effectiveness of Thinking in Different Tasks.} In this section, we investigate the role of thinking in different tasks during RL training. By default, we apply the think-before-answer format to the chart question answering task. We further explore the impact of extending this format to spatial and textual learning. The results, shown in Table~\ref{tab:ablation_on_thinking}, demonstrate that incorporating thinking consistently improves performance across benchmarks. Notably, even our grounding benchmark (CS-Bench) benefits from this approach. This finding aligns with prior works~\cite{wang2025time,li2025videochat}, suggesting that low-level visual understanding can also be enhanced by thinking, as it provides richer context to the MLLM before producing the final answer.

\subsection{Analysis and Visualization.} \label{analysis_and_visualization}
% \noindent\textbf{Data Distribution Analysis of the Chart Images}. Figure~\ref{fig:visualization_tsne_and_grounding}-A presents a t-SNE embedding of SigLIP~\cite{zhai2023sigmoid} features across four chart datasets. CharXiv—sourced from arXiv papers—reflects high-quality, real-world charts with complex visual structure and dense details. In contrast, ReachQA and ChartQA concentrate into a few tight clusters that are displaced from the CharXiv region, indicating both a distribution shift and limited visual variety relative to real-world charts. Our dataset largely overlaps the CharXiv's region yet extends beyond it with denser coverage and a broader spread across categories, suggesting a closer match to real-world distribution and greater intra-class diversity.

% \noindent\textbf{Grounding Dataset Visualization.} Figure~\ref{fig:visualization_tsne_and_grounding}-B shows the chart element bounding boxes in our grounding dataset. Our LLM code-evolution-based method can be effectively applied into different chart type and diverse subplot layout, finding the chart element locations accurately. 

% \noindent\textbf{Benefit of Spatial-textual Learning.} 
Figure~\ref{fig:qa_code_grounding} shows the prediction from START and Qwen2.5-VL on different tasks. By explicitly learning the chart element grounding, the spatial understanding of the model is enhanced, thus fixing the error caused by the grounding wrong chart subplot in the chart question answering setting (Subplot A). It also enhances the bounding box location prediction (Subplot B). With the chart-to-code learning, we improve the textual understanding of the model, which is reflected by a better re-rendered chart image (Subplot C).

\section{Conclusion and Discussion}
\label{sec:conclusion}
% need to update the code
In this paper, we present START, a spatial–textual learning framework for chart understanding that reflects the dual nature of charts—their structured visual layout and their underlying textual content. START couples chart-element grounding with chart-to-code to jointly learn spatial structure and textual details on the chart. To support this, we introduce START-Dataset with a data-construction pipeline that first transcribes real-world charts into Python plotting code using a MLLM, then evolves the code with an LLM to obtain the precise chart-element locations. For evaluating the MLLM's spatial understanding of the Chart, we propose the Chart Spatial understanding Benchmark (CS-Bench) to support the comprehensive chart understanding evaluation. Benefiting from this spatial–textual supervision, START achieves substantial improvements on benchmarks over strong baselines and surpasses the previous state-of-the-art by a clear margin. We hope our work shed light on the way to achieving deeper chart intelligence. 

\medskip
\noindent \textbf{Acknowledgment}: We thank Professor Yin Li, Dr. Yuhua Chen, and Pengfei Yu for their helpful discussions and valuable suggestions throughout this project. We also appreciate AWS for providing compute and API resources.
{
    \small
    \bibliographystyle{ieeenat_fullname}
    \bibliography{main}
}
% WARNING: do not forget to delete the supplementary pages from your submission 
\clearpage
% \FloatBarrier
\appendix
In this supplement, we (1) show additional START-dataset construction details and visualization in (Section~\ref{dataset_supple}) (2) present additional details for the Chart Spatial understanding Benchmark (CS-Bench) in (Section~\ref{benchmark_supple}) (3) describe additional implementation details and results in (Section~\ref{implementation_exp_supple}).
We hope that this document will complement our main paper.

\section{START-Dataset}\label{dataset_supple}
In this section, we provide additional details in the dataset construction pipeline.
\subsection{Chart-to-code}\label{chart_to_code_supple}
\begin{figure}[t!]
    \centering
    \includegraphics[width=0.95\linewidth]{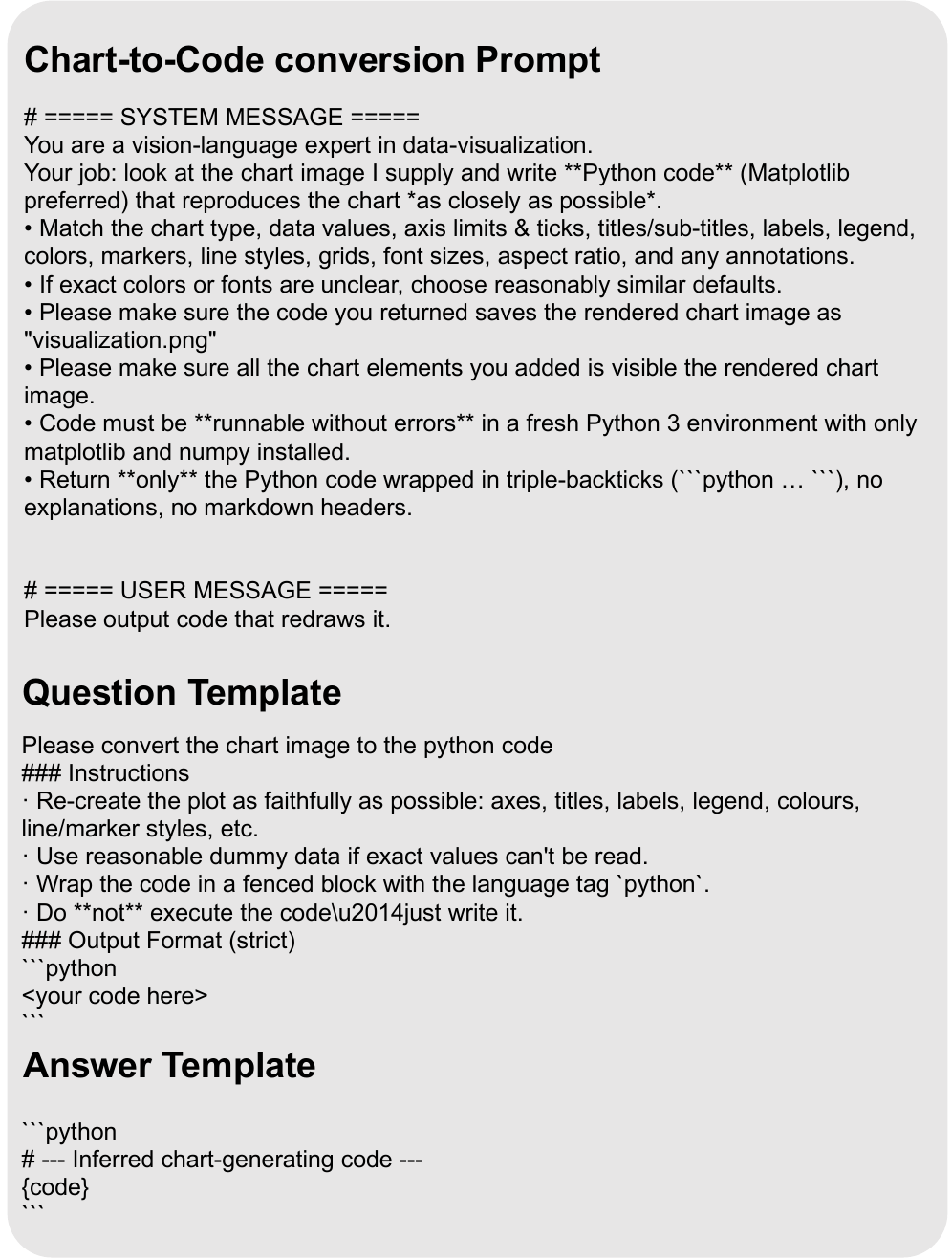}
    \vspace{-0.5em}
    \caption{The prompt we use for converting chart to code, and the template we use for preparing the annotations in $D_c$.}
    \label{fig:chart_to_code_annotation_prompt}
    \vspace{-0.5em}
\end{figure}

For converting the chart images to Python code during the dataset construction, we explore different approaches: 1. Directly use the multi-modal large language model (MLLM) to convert the chart image to Python code. We tried Qwen2.5-VL~\cite{bai2025qwen2} and a proprietary model. 2. Use the MLLM as a chart captioner to convert the chart image to a chart description first and then use the Large Language Model (LLM) as the coder to generate Python code based on the chart description. We tried Qwen2.5-VL as the captioner and used an open-source LLM as the coder, in the hope that applying the captioner first before creating the Python code can preserve more chart details.
The visualization in Figure~\ref{fig:different_chart_to_code} shows that directly using a proprietary model could produce the most authentic reproduced chart images and preserve most details on the original charts. 
We share the prompt we use to convert a chart image to code with the proprietary model in Figure~\ref{fig:chart_to_code_annotation_prompt}.

To construct the chart-to-code dataset $D_c$, we first use the proprietary model to filter the non-chart images in the ArxivQA~\cite{li2024multimodal}. After obtaining the chart images, we prompt the proprietary model to convert the chart image to Python code. We run the Python code and generate the reproduced chart images. We then filter the distorted reproduced chart images by prompting the proprietary model.
After we obtain the Python code generated by the proprietary model and the reproduced chart images, we construct the dataset that is used in textual learning during the supervised finetuning (SFT) training and the reinforcement learning (RL). 
For the SFT, we use the fixed templates to construct the question and answer pair. Please see the templates in figure~\ref{fig:chart_to_code_annotation_prompt}.
For the dataset used in RL, we use a fixed prompt to ask the model to convert the chart image to the code. We use the code generated by the proprietary model as the Ground Truth and feed it to the grader to calculate the reward for the model prediction.

% \begin{figure}[t!]
%     \centering
%     \includegraphics[width=0.95\linewidth]{figures/find_chart_element_location_code_snippet.png}
%     \vspace{-0.5em}
%     \caption{Code snippet about how we find the chart element location in the sample evolved Python code.}
%     \label{fig:find_chart_element_location_code_snippet}
%     \vspace{-0.5em}
% \end{figure}

\begin{algorithm}
\caption{Extracting Chart Elements’ Location}
\label{alg:extract_plot}
\begin{lstlisting}[language=Python, basicstyle=\ttfamily\small, breaklines=true]
# Get title location
title_obj = ax.title
if title_obj.get_text():
    subplot_data['title'] = get_bbox_pixels_flipped(title_obj, renderer, img_height)

# Get x-axis label location and name
xlabel_obj = ax.xaxis.label
x_axis_name = xlabel_obj.get_text() if xlabel_obj.get_text() else 'x-axis'
if xlabel_obj.get_text():
    subplot_data['x_axis_names'].append(get_bbox_pixels_flipped(xlabel_obj, renderer, img_height))

# Get y-axis label location and name
ylabel_obj = ax.yaxis.label
y_axis_name = ylabel_obj.get_text() if ylabel_obj.get_text() else 'y-axis'
if ylabel_obj.get_text():
    subplot_data['y_axis_names'].append(get_bbox_pixels_flipped(ylabel_obj, renderer, img_height))

# Get x-axis tick locations - filter for valid ticks
x_ticks = []
valid_ticks = [str(decade) for decade in decades]
for tick in ax.get_xticklabels():
    if tick.get_visible() and tick.get_text().strip() in valid_ticks:
        x_ticks.append(get_bbox_pixels_flipped(tick, renderer, img_height))
if x_ticks:
    subplot_data['x_axis_ticks'][x_axis_name] = x_ticks

\end{lstlisting}
\end{algorithm}

\begin{figure}[t!]
    \centering
    \includegraphics[width=0.95\linewidth]{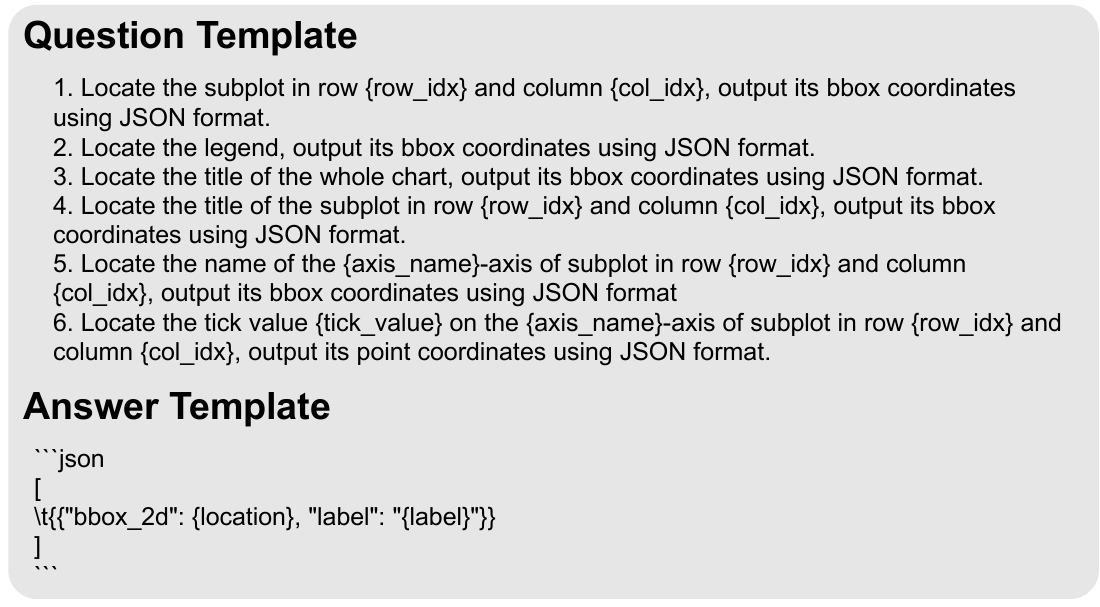}
    \vspace{-0.5em}
    \caption{Chart element location annotation $D_g$ preparation template.}
    \label{fig:chart_element_location_annotation_template}
    \vspace{-0.5em}
\end{figure}

\subsection{Location Generation} \label{location_gen}

To construct the chart element grounding dataset $D_g$, we first generate sample evolved code for each chart category. These evolved codes utilize Matplotlib's built-in functions to automatically extract the locations of chart elements from the rendered chart image. The extracted locations are then saved in JSON format.
Algorithm~\ref{alg:extract_plot} provides a code snippet illustrating this process.
By providing these standardized sample codes as examples during the LLM-driven code evolution process, we both improve the success rate of generating accurate code and ensure uniformity in how chart element locations are stored in JSON file.

We use these sample evolved codes as examples to prompt the proprietary model to evolve the Python codes, which we obtained from the chart-to-code process.
The prompt we use for code evolution is provided in Figure~\ref{fig:chart_evolve_prompt}.
We then execute the evolved code to obtain the locations of the chart elements. 
Next, we uniformly sample these locations for each type of chart element. Finally, we convert the sampled locations into grounding annotations by applying fixed templates.
Please see the template details in Figure~\ref{fig:chart_element_location_annotation_template}.

% TODO: HOW to construct the $D_g$
% TODO: The prompt to convert the location to qa annotation (the fixed prompt)

\begin{figure*}[t!]
    \centering
    \includegraphics[width=0.95\linewidth]{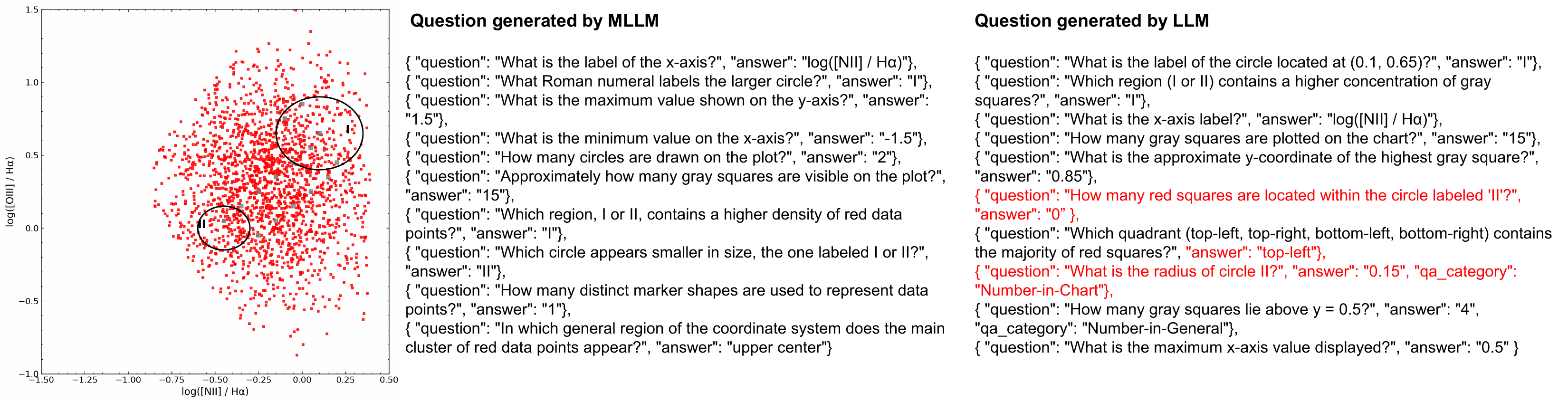}
    \vspace{-0.5em}
    \caption{The QA samples generated by the MLLM and the LLM. We highlight the questions that are improperly hard or the incorrect answers in red.}
    \label{fig:supplment_qa_samples}
    \vspace{-0.5em}
\end{figure*}

\begin{figure*}[t!]
    \centering
    \includegraphics[width=0.95\linewidth]{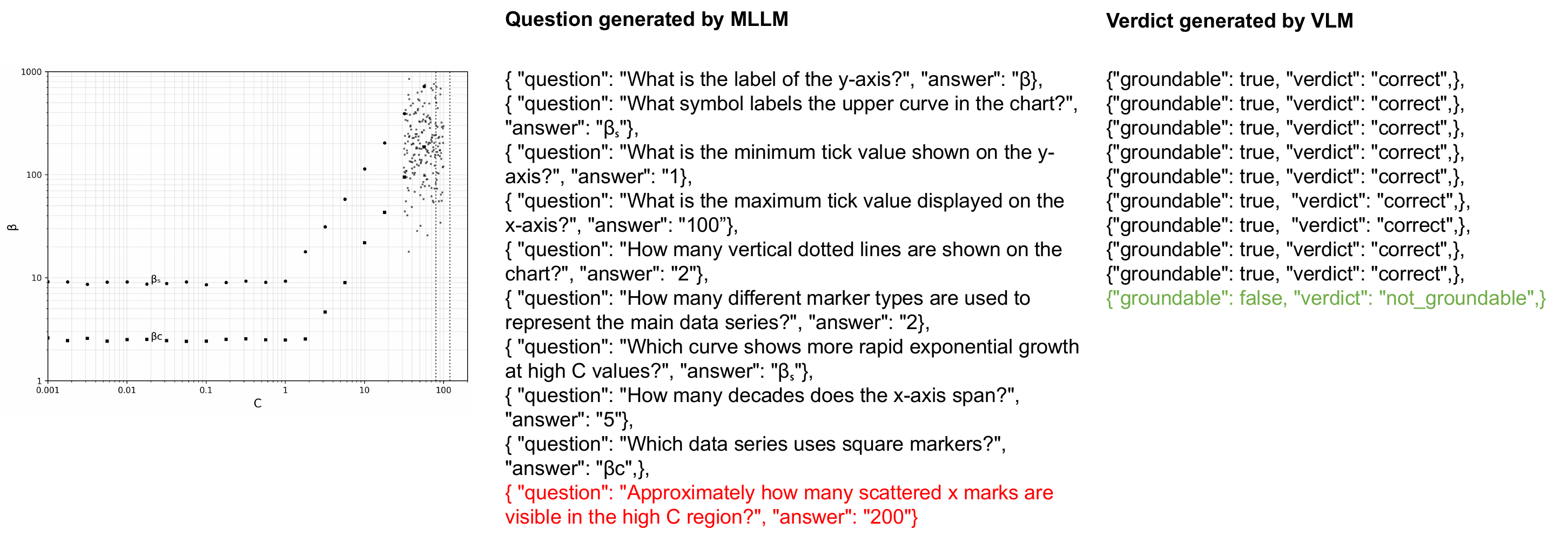}
    \vspace{-0.5em}
    \caption{The question-answer pairs and the corresponding verdicts. We highlight the questions that are improperly in red and the corresponding verdicts in green.}
    \label{fig:supplment_verdict_samples}
    \vspace{-0.5em}
\end{figure*}

\subsection{QA generation}
For generating the QA pair for the rendered chart images, we try two different ways. 
1. Use a MLLM with a chart image and the corresponding Python code to generate the question and answer pair.
2. Use an LLM, which takes the Python code as input to generate the question and answer pair.
The MLLM approach leverages both visual and code inputs, enabling it to generate questions grounded in spatial properties of the chart (e.g., the spatial arrangement of scatter points). In contrast, the LLM approach focuses solely on the Python code, making it well-suited for capturing fine-grained chart details, such as the exact number of points plotted or specific data values. Figure~\ref{fig:supplment_qa_samples} shows a visualization of sample questions generated by both MLLM and LLM.
In our experiments, we found that MLLM consistently produced higher-quality QA pairs. We thus use MLLM to generate our question-answer pairs.

We curated a set of high-quality examples and used them as part of a few-shot prompt to guide MLLM in generating ten QA pairs for each chart image. The full prompt used for this generation process is provided in Figure~\ref{fig:qa_generation_prompt}.

To ensure quality, we incorporate a verification step to detect and remove unreasonable questions or incorrect answers. Specifically, we prompt a strong MLLM to assess whether each question is groundable (i.e., tied to elements visible in the chart) and answerable (i.e., solvable by a MLLM). We filter out hallucinated questions that reference non-existent elements or those beyond the MLLM’s capacity (e.g., precisely counting 200 dots). In addition, the MLLM verifies the correctness of each answer. Based on these verdicts, we filter the QA pairs to obtain the final dataset for the chart question answering task, denoted as $D_q$.
We present some verdict samples given by the strong MLLM for the QA pair generated by MLLM in Figure~\ref{fig:supplment_verdict_samples}. 

\subsection{Curate the SFT and RL data splits}
We combine the data from chart question answering, spatial learning, and textual learning as the Supervised-Finetuning Dataset (START-Dataset-SFT), and we sample the Reinforcement Learning Dataset (START-Dataset-RL) from the START-Dataset-SFT based on the question difficulties. To determine the difficulty of the training samples, We run the Qwen2.5-VL on the training set and regard the probability of the model can give the correct answer to a question as difficulty. We use the difficulty as weight for sampling and get the START-Dataset-RL.

\begin{figure*}[t!]
    \centering
    \includegraphics[width=0.95\linewidth]{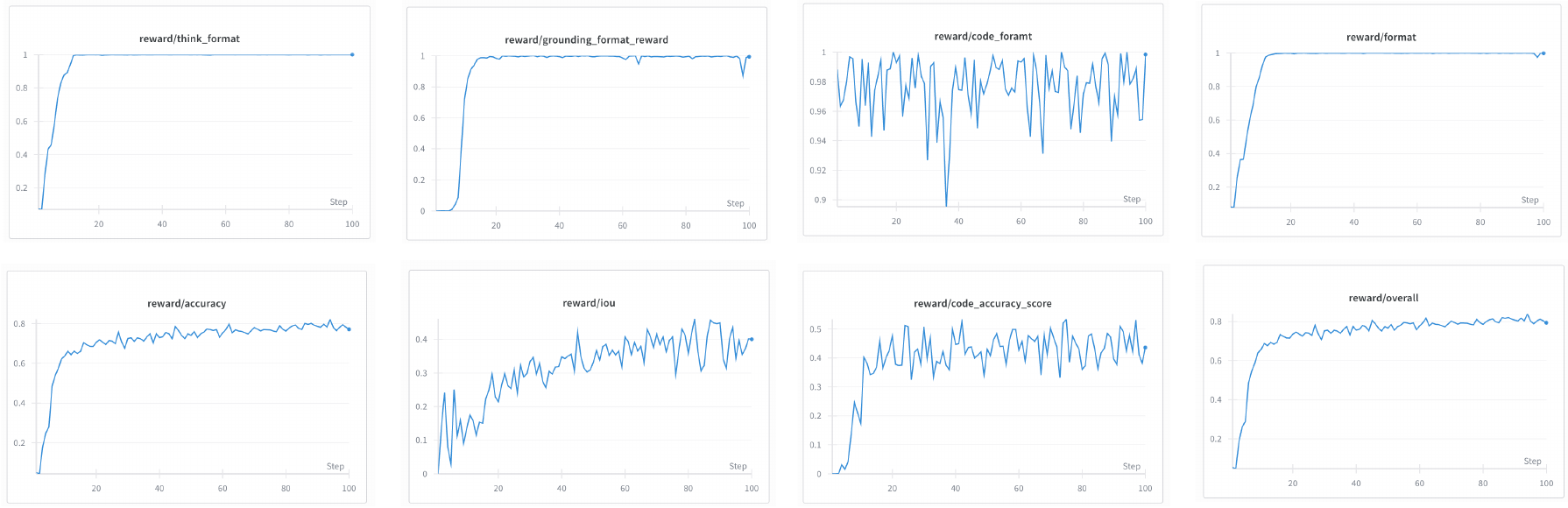}
    \vspace{-0.5em}
    \caption{learning curve of START-RL-7B, trained with chart question answering, chart element grounding, and the chart-to-code task.}
    \label{fig:learning_curve}
    \vspace{-0.5em}
\end{figure*}

\section{CS-Bench}\label{benchmark_supple}
In this section, we provide more details on the construction pipeline for Chart Spatial understanding Benchmark (CS-Bench). 
% The benchmark statistic is in Figure~\ref{fig:benchmark_stat}.

\noindent\textbf{Images.} 
The benchmark consists of 613 images rendered from a holdout subset of code evolved during the START-Dataset construction. 
% Its image distribution, shown in 'Chart Image' part of Figure~\ref{fig:benchmark_stat}, 

\noindent\textbf{Chart Element Locations.}
Given CS-Bench is built based on the evolved Python code, the JSON file that stores the location of the chart elements will be generated when the Python code is run.

\noindent\textbf{The Question Answer pairs.}
Our benchmark features two primary types of inquiries: grounding questions and QA grounding questions.
A grounding question directly prompts a model to find the location of a specific element within a chart. In contrast, a QA grounding question presents a two-part task: it first asks a question related to the chart's content and then requires the model to identify the location of the chart element(s) referenced in the answer or in the question.
CS-Bench has 350 grounding questions and 342 QA grounding questions. 
For the grounding question, we following the same pipeline used in generating the chart element location dataset $D_g$ in section~\ref{location_gen} to generate the question answer pairs.
For the QA grounding question, we prompt a MLLM to generate the question related to the chart image, and then we pick the location mentioned in the question or the answer. 
The question and the answer are manually verified.
% The question and the location distribution are shown in 'Chart QA' and 'Grounding Data' Parts in Figure~\ref{fig:benchmark_stat}.
Sample chart images and the questions could be found in Figure~\ref{fig:benchmark_sample_visualization}.

\noindent\textbf{The Evaluation Metrics.}
Considering the Ground Truth location has a lot of small items, for instance, ticks values or axis names, we use recall at IoU 0.3 (recall@0.3) as the main metric of this benchmark. We report the recall@0.3 for 692 ground truth bounding boxes. We also present the accuracy of answer to the 342 QA grounding questions as an auxiliary metric.

\section{START Experiments}\label{implementation_exp_supple}
\subsection{Benchmarks.}
We use CharXiv~\cite{wang2024charxiv} validation split, ChartQA~\cite{masry2022chartqa} test split, ChartQAPro~\cite{masry2025chartqapro}, ChartMimic~\cite{yangchartmimic} and CS-Bench proposed by us as benchmarks. 
Specifically, CharXiv has 1000 chart images which come from arXiv papers, and it splits the evaluation set into 4k descriptive questions and 1k reasoning questions. While the descriptive questions focus on the general chart elements, for instance, title, legend, and tick values on the axis, the reasoning questions focus on the trend or conclusion reflected from the chart.
It use GPT-4o~\cite{hurst2024gpt} as judge and adapt accuracy as metric.
ChartQA contains 1509 images and 2500 questions, which mostly focus on line, bar, and pie charts. It uses accuracy as a metric.
ChartQAPro consists of 1341 chart images, which come from 157 diverse online platforms and are paired with 1948 questions, which are divided into factoid, multiple-choice, conversational, hypothetical, and fact-checking. It uses accuracy as a metric.
ChartMimic contains 600 human-curated (figure, instruction, code) triplets and evaluates the model's ability to convert a chart image to code. 
It use GPT-4o as the judge and adapts the score as metric. CS-Bench evaluates the spatial understanding of the MLLM toward the chart and uses recall at IoU 0.3 (recall@0.3) of the GT bounding box and accuracy of the answer to the question as metrics. We regard recall@0.3 as the main metric for CS-Bench.

\subsection{Implementation details.}

\textbf{Training details in Supervised Finetuning (SFT).}

\textbf{\textit{Initialization.}} 
We initiate the Supervised Finetuning (SFT) training with Qwen2.5-VL-3B and 7B model~\cite{bai2025qwen2} checkpoints, given that they have promising reasoning ability that could be elicited during the RL training~\cite{huang2025vision, yang2025r1, feng2025video} and reasonable grounding performance, thanks to its high-quality pretraining. 
% We use LLaMA-Factory~\cite{zheng2024llamafactory} as the codebase.

\textbf{\textit{Training Hyper-parameters.}}
We train the model with learning 1e-6 with a 0.1 warm-up ratio and cosine learning rate decay. We set the global batch size to 128. We train the models with SFT on the START-SFT dataset for 1 epoch.

\textbf{\textit{Data.}} 
For different SFT settings, we mix different portions of the START-SFT as the training dataset. For instance, if we use SFT to train the model with chart question answering and chart element grounding, we mix these two portions of the data in the START-SFT as training data to train the model.

\noindent\textbf{Training details in Reinforcement Learning (RL).}

\textbf{\textit{Initialization.}}
To start the RL training, we initialize the model with the corresponding SFT checkpoint. For instance, when we conduct the RL training with chart question answering and chart element grounding, we use the SFT checkpoint, which is also trained with the chart question answering and chart element grounding tasks to initialize the model. 
%For the RL training, we use the Easy-R1~\cite{zheng2025easyr1} as the codebase. 

\textbf{\textit{Training Hyper-parameters.}}
We train the model for 100 steps with a learning rate of 1e-6, rollout batch size 512, and rollout number of 5. During the training, we set the micro-batch size per device to 4. We train all the models with 8 A100 GPUs.

\textbf{\textit{Data.}}
Similar to the setting in the SFT, we mix different portions of the START-RL dataset as training dataset. For instance, if we train the model with RL with chart QA and chart element grounding, we will use the question from the chart question answering and chart element grounding split of START-RL to prompt the model rollout the answer during the training. We use the ground truth answer or the bounding box as the reference to calculate the reward for the response. 

\textbf{\textit{Reward design and reward calculation.}}
For the reward calculation, we consider 2 different types of rewards, the formatting reward and the accuracy reward. For the formatting reward, we mainly use a regular expression to judge whether the model's answer fits into a required format. The value of the reward is either 0 or 1. The details is included in Figure 2 of the main paper. 
% For the accuracy reward, we use the grade\_answer function from the Mathruler~\cite{mathruler} to judge the answer, which will give the reward value either 1 or 0. 
For the chart element grounding task, we use the IoU value between the predicted bounding box and the ground truth bounding box as the reward, the value will be a float value falls between 0 to 1. For the chart-to-code task, we use an LLM to judge the predicted code from different perspectives with the Ground Truth code as a reference. The prompt we use is shown in Figure~\ref{fig:code_judge_reward_prompt}. 
We judge the code from five different perspectives, which include data, plot type structure, axes scales and limits, text elements, and styling. Each element will be graded with a score of 0 to 5. We then sum up the score from different perspectives and normalize the score between 0 to 1 as the reward.

\textbf{\textit{The training curve.}} Figure~\ref{fig:learning_curve} shows the learning curve of START-RL-7B, trained with chart question answering, chart element grounding, and the chart-to-code task. It shows the format and accuracy reward for chart question answering (think\_format, accuracy), the chart element grounding (grounding\_format\_reward, iou), and chart-to-code (code\_format, code\_accuracy\_score). It also includes the overall format and overall reward (format, and overall).

% 2. The ablation study for the code and location.?? (any other ablation, the dataset distribution, more grounding, more code?)

\begin{figure*}[t!]
    \centering
    \includegraphics[width=0.95\linewidth]{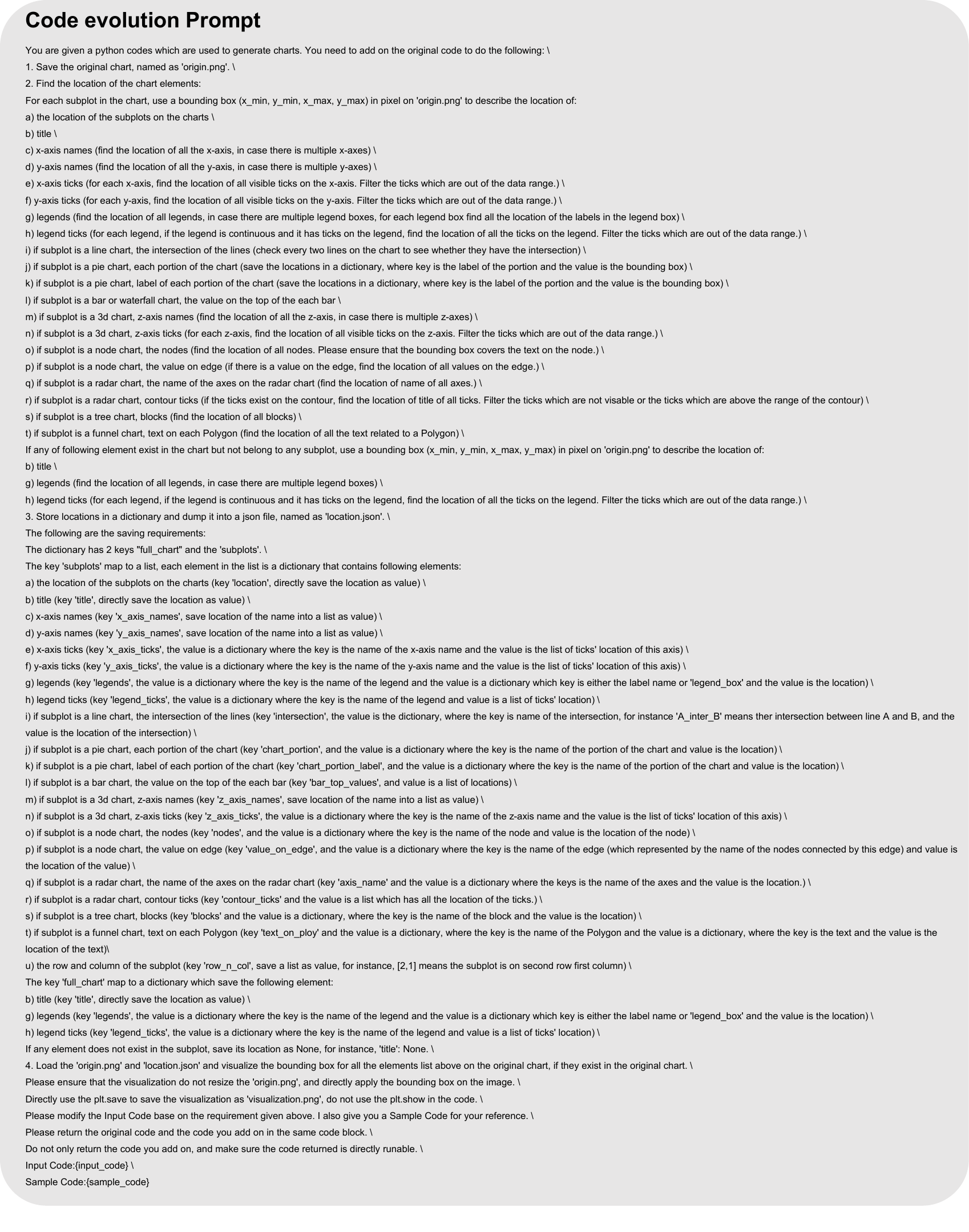}
    \vspace{-0.5em}
    \caption{Prompt for code evolution.}
    \label{fig:chart_evolve_prompt}
    \vspace{-1.5em}
\end{figure*}

\begin{figure*}[t!]
    \centering
    \includegraphics[width=0.9\linewidth]{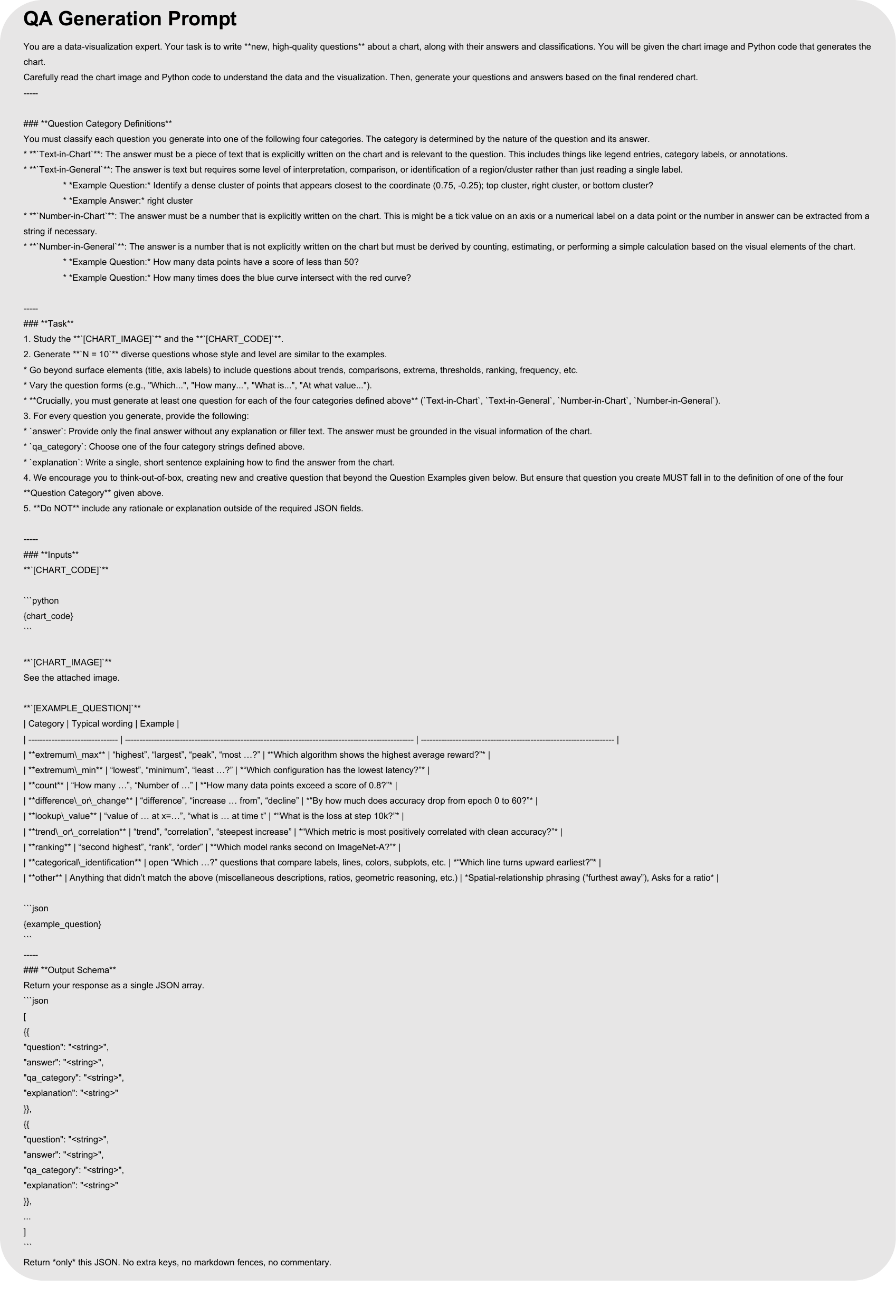}
    \vspace{-2em}
    \caption{Prompt for QA generation.}
    \label{fig:qa_generation_prompt}
    \vspace{-1.5em}
\end{figure*}

\begin{figure*}[t!]
    \centering
    \includegraphics[width=0.9\linewidth]{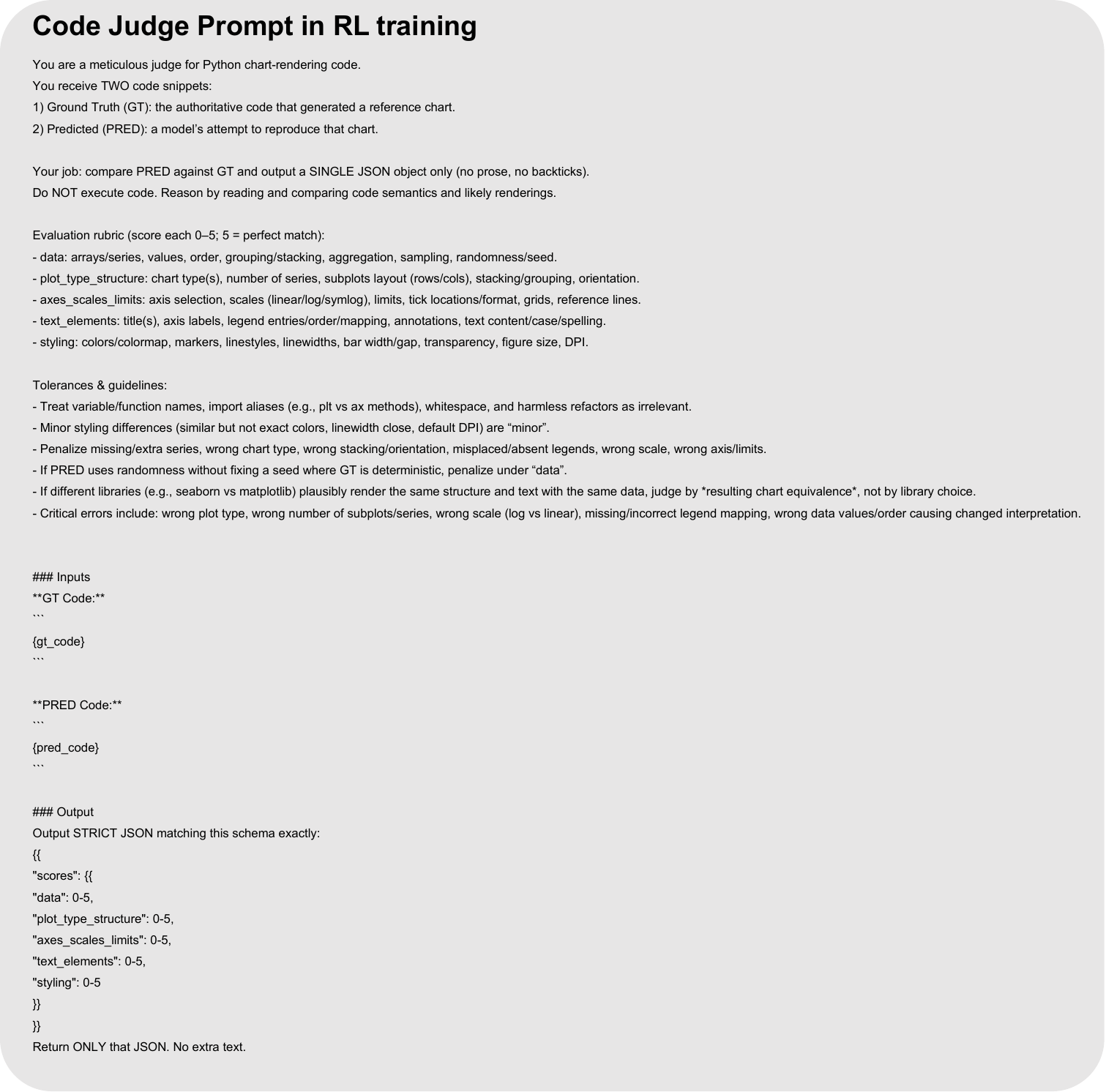}
    \vspace{-0.5em}
    \caption{Prompt for code generation during the RL training.}
    \label{fig:code_judge_reward_prompt}
    \vspace{-1.5em}
\end{figure*}

\begin{figure*}[t!]
    \centering
    \includegraphics[width=1.0\linewidth]{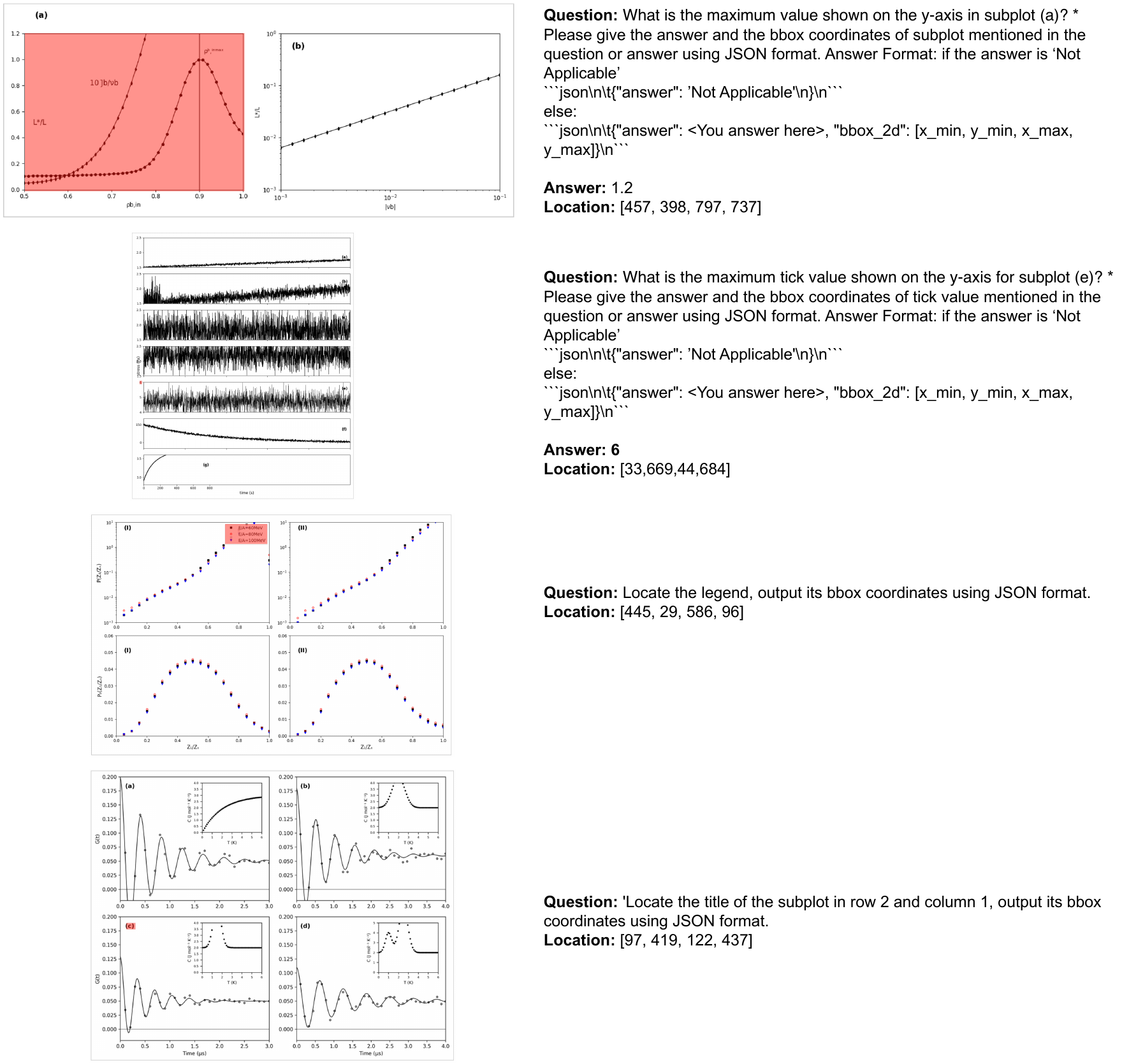}
    \vspace{-0.5em}
    \caption{The visualization of the samples in the Chart Spatial understanding Benchmark (CS-Bench). We visualize the bounding box region in red.}
    \label{fig:benchmark_sample_visualization}
    \vspace{-1.5em}
\end{figure*}

\begin{figure*}[t!]
    \centering
    \includegraphics[width=0.8\linewidth]{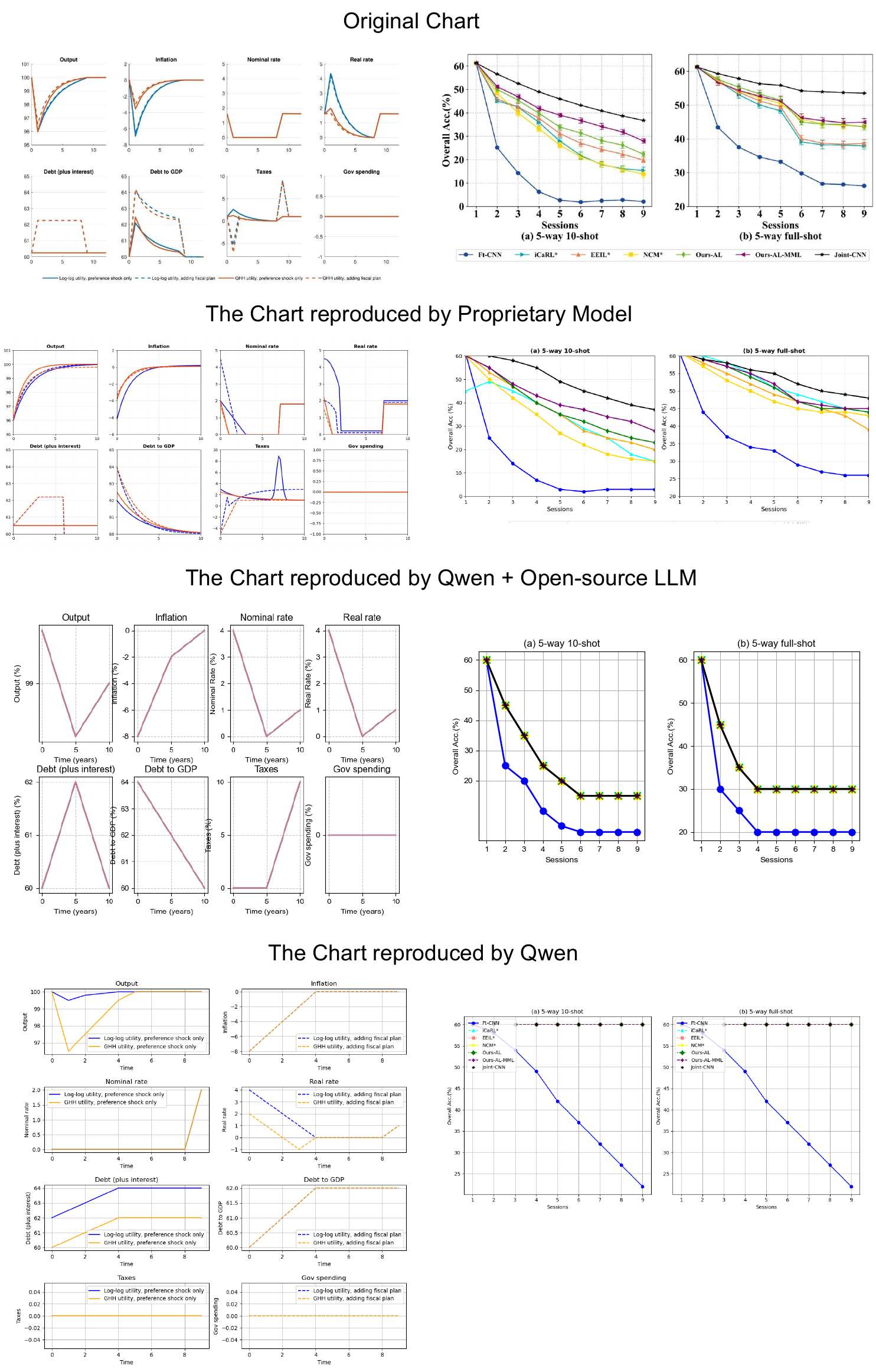}
    \vspace{-1em}
    \caption{Visualization of reproduced charts using different chart-to-code methods.}
    \label{fig:different_chart_to_code}
    \vspace{-0.5em}
\end{figure*}

\end{document}